%% file: klog5.tex
\documentclass[preprint,12pt]{elsarticle} 
\usepackage[T1]{fontenc}

\newif\ifarxiv
\arxivtrue

\usepackage{graphicx}
\usepackage{array}
\usepackage{amsmath}
\usepackage{amsfonts}
\usepackage{amssymb}

\usepackage{mathrsfs}

\usepackage{multirow}
\usepackage[draft]{fixme}

\usepackage{fancyvrb}
\usepackage{bnf}
\usepackage{url}
\usepackage[retainorgcmds]{IEEEtrantools}

\usepackage{float}
\floatstyle{plain}
\newfloat{listing}{thp}{section}
\floatname{listing}{Listing}

\newtheorem{example}{Example}[section]
\let\olddexample\example
\renewcommand{\example}{\olddexample\normalfont}

\newcommand*{\argmax}{\textnormal{arg}\max}
\newcommand*{\RSet}[0]{\ensuremath{\mathbb{R}}}

\newcommand*{\CalN}{\ensuremath{\mathcal{N}}}
\newcommand*{\CalL}{\ensuremath{\mathcal{L}}}
\newcommand*{\CalD}{\ensuremath{\mathcal{D}}}

\newcommand*{\IN}[0]{\ensuremath{\mathbb{N}}}

\newcommand{\Label}{\CalL}
\newcommand{\transpose}{^{\top}}
\newcommand{\maxradius}{\ensuremath{r^*}}
\newcommand{\maxdistance}{\ensuremath{d^*}}

\makeatletter
\def\namedlabel#1#2{#2\begingroup
   \def\@currentlabel{#2}%
   \label{#1}\endgroup
}
\makeatother

\journal{Artificial Intelligence}

\begin{document}

\begin{frontmatter}

\title{kLog: A Language for Logical and  Relational Learning with Kernels}
\tnotetext[label1]{PF was a visiting professor at KU Leuven and FC a postdoctoral fellow at KU Leuven while this work was initiated}
\author[unifi,kuleuven]{Paolo Frasconi\corref{cor1}}
\ead{p-f@dsi.unifi.it}
\author[freiburg,kuleuven]{Fabrizio Costa}
\ead{costa@informatik.uni-freiburg.de}
\author[kuleuven]{Luc De Raedt}
\ead{Luc.DeRaedt@cs.kuleuven.be}
\author[kuleuven]{Kurt De Grave}
\ead{Kurt.DeGrave@cs.kuleuven.be}
\cortext[cor1]{Corresponding author}

\address[unifi]{Dipartimento di Ingegneria dell'Informazione,
Universit{\`a} degli Studi di Firenze, via di Santa Marta 3, I-50139 Firenze, Italy}
\address[freiburg]{Institut f\"ur Informatik, Albert-Ludwigs-Universit\"at, Georges-Koehler-Allee, Geb 106, D-79110 Freiburg, Germany}
\address[kuleuven]{Departement Computerwetenschappen, 
KU Leuven, Celestijnenlaan 200A, B-3001~Heverlee, Belgium}

\begin{abstract}
  We introduce kLog, a novel approach to statistical relational learning.
  Unlike standard approaches, kLog does not 
  represent a probability distribution directly.  It is rather 
  a language to perform \emph{kernel-based learning} on expressive logical and relational representations.
  kLog allows users to specify learning problems declaratively.
  It builds on simple but powerful concepts: learning from
  interpretations, entity/rela\-tionship data modeling, logic
  programming, and deductive databases. 
  Access by the kernel to the rich representation is mediated by a technique we call \emph{graphicalization}: 
  the relational representation is first transformed into a graph --- in particular, a grounded entity/relationship diagram. 
  Subsequently, a choice of graph kernel defines the feature space.
  kLog supports mixed numerical and symbolic data, as well as
  background knowledge in the form of Prolog or Datalog programs as
  in inductive logic programming systems.
  The kLog framework can be applied to tackle the same range
  of tasks that has made statistical relational learning so popular,
  including classification, regression, multitask learning, and
  collective classification. 
  We also report about empirical comparisons, showing that kLog can be
  either more accurate, or much faster at the same level of accuracy,
  than Tilde and Alchemy.
  kLog is GPLv3 licensed and is available at \url{http://klog.dinfo.unifi.it} along with tutorials.
\end{abstract}

\end{frontmatter}

\section{Introduction}

The field of statistical relational learning (SRL) is populated with a
fairly large number of models and alternative representations, a
state of affairs often referred to as the ``SRL alphabet
soup''~\cite{Dietterich:2008:Structured-machine-learning:,De-Raedt:2008:Towards-digesting-the-alphabet-soup}.
Even though there are many differences between these approaches, they
typically extend a probabilistic representation (most often, a
graphical model) with a logical or relational one~\cite{:2008:Probabilistic-inductive-logic,Getoor:2007:Introduction-to-statistical-relational}.
The resulting models then define a probability distribution over
possible worlds, which are typically (Herbrand) interpretations
assigning a truth value to every ground fact.

However, the machine learning literature 
contains --- in addition to probabilistic graphical models --- several other types
of statistical learning methods. In particular, kernel-based learning and support vector machines
are amongst the most popular and powerful machine learning systems that exist today. 
But this type of learning system has --- with a few notable exceptions to relational prediction~\cite{Landwehr:2010:Fast-learning-of-relational,Taskar:2003:Max-margin-Markov-networks}  --- 
not yet received a lot of attention in the SRL literature. 
 Furthermore, while it is by now commonly accepted that frameworks like Markov logic networks (MLNs)~\cite{Richardson:2006:Markov-logic-networks}, probabilistic relational models (PRMs)~\cite{friedman99:learn}, or probabilistic programming~\cite{:2008:Probabilistic-inductive-logic,Getoor:2007:Introduction-to-statistical-relational}
are general logical and relational languages that support a wide range of learning tasks, there exists today no such language 
for kernel-based learning.  It is precisely this gap that we ultimately want to fill.   


This paper introduces the kLog language and framework for kernel-based logical and relational learning.
The key contributions of this framework are threefold: 1) kLog is a language that allows users to declaratively specify relational learning tasks in a similar way as statistical relational learning and inductive logic programming approaches but it is based 
on kernel methods rather than on probabilistic modeling; 2) kLog compiles the relational domain and learning task into a graph-based representation using a technique called graphicalization; and 3) kLog uses a graph kernel to construct the feature 
space where eventually the learning takes place. This whole process is reminiscent of knowledge-based model construction in statistical relational learning.  We now sketch these contributions in more detail and discuss the relationships with
statistical relational learning.

One contribution of this paper is the introduction of the kLog language and framework for kernel-based logical and relational learning.  
kLog is embedded in Prolog (hence the name) and
allows users to specify different types of logical and relational learning problems at a high level in a declarative way.  
kLog adopts, as  many other logical and relational learning systems, the learning from interpretations framework~\cite{De-Raedt:2008:Logical-and-relational-learning}. In this way,
the entities (or objects) and the 
relationships amongst them can be naturally represented.  However, unlike typical statistical relational
learning frameworks, kLog does not employ a probabilistic framework
but is rather based on linear modeling in a kernel-defined feature
space. 

kLog constitutes a first step into the direction of a general kernel-based SRL approach.
kLog generates a set of features starting from a logical and
relational learning problem and uses these features for learning a  (linear) statistical model.
This is not unlike Markov logic but there are two major differences. First, kLog is based on a linear statistical model instead of a Markov network. Second, the feature space is not immediately defined by the declared logical formulae but it is constructed by a graph kernel, where nodes in the graph correspond to entities and relations, some given in the data, and some (expressing background knowledge) defined declaratively in Prolog. Complexity of logical formulae being comparable, graph kernels leverage a much richer feature space than MLNs. 
In order to learn, kLog essentially describes learning problems at three different levels.  The first
level specifies the logical and relational learning problem. At this
level, the description consists of an E/R-model describing the structure of the data and the data itself,  
which is similar to that of traditional SRL systems~\cite{Heckerman:2007:SRL}. 
The data at this level is then \emph{graphicalized}, that is, the interpretations are
transformed into graphs.  This leads to the specification of a graph
learning problem at the second level. Graphicalization is the equivalent of knowledge-based model construction.
Indeed, SRL systems such as PRMs and MLNs also (at least conceptually) produce graphs, although these graphs
represent probabilistic graphical models.  Finally, the graphs produced by kLog are turned
into feature vectors using a graph kernel, which leads to a statistical learning problem at the third level.  
Again there is an analogy with systems such as Markov logic as the Markov network that is generated 
in knowledge-based model construction lists also the features.  Like in these systems, kLog features are tied together as every occurrence 
is counted and is captured by a single same parameter in the final linear model.  

It is important to realize that kLog is a very flexible architecture
in which only the specification language of the first level is fixed;
at this level, we employ an entity/relationship (E/R) model. The
second level is then completely determined by the choice of the graph
kernel. In the current implementation of kLog that we describe in this
paper, we employ the neighborhood subgraph pairwise distance kernel
(NSPDK) \cite{Costa::Fast-neighborhood-subgraph} but the reader
should keep in mind that other graph kernels can be incorporated.
Similarly, for learning the linear model at level three we mainly
experimented with variants of SVM learners but again it is important
to realize that other learners can be plugged in.  This situation is
akin to that for other statistical relational learning representations, 
for which a multitude of
different inference and learning algorithms has been devised (see also
Section~\ref{sec:related} for a discussion of the relationships
between kLog and other SRL systems).

In the next section, we illustrate the main steps of kLog modeling with a complete example in a real world domain.
In Section~\ref{sec:statistical-modeling}, we formalize the assumed statistical setting for supervised learning from interpretations, 
provide some background on statistical modeling from a relational learning point of view, and position kLog more clearly in the context of related systems such as Markov logic, M$^3$N~\cite{Taskar:2003:Max-margin-Markov-networks}, etc. 
In Section~\ref{sec:klog-language}, we formalize the semantics of the language and illustrate what types of learning problems can be formulated in the framework. Further examples are given in Section~\ref{sec:examples}. The graphicalization approach and the graph kernel are detailed in Section~\ref{sec:graphicalization}. Some empirical evaluation is reported in Section~\ref{sec:klog-in-practice} and, finally, the relationships to other SRL systems are discussed in Section~\ref{sec:related}.

\section{A kLog example}
\label{sec:running-example}
Before delving into the technical details of kLog, we illustrate the different steps
on a real-life example using the UW-CSE dataset prepared by Domingos et al.\ for  
demonstrating the capabilities of MLNs~\cite{Richardson:2006:Markov-logic-networks}. Anonymous data was obtained from the University of Washington Department of Computer Science and Engineering. Basic entities include persons (students or professors), scientific papers, and academic courses. Available relations specify, e.g., whether a person was the author of a certain paper, or whether he/she participated in the teaching activities of a certain course. The learning task consists of predicting students' advisors, i.e., to predict the binary relation \textsf{advised\_by} between students and professors.

\subsection{Data format}
Data comes in the form of true ground atoms, under the closed-world assumption. Since (first-order logic) functions are not allowed in the language, a ground atom is essentially like a tuple in a relational database, for example 
\textsf{taught\_by(course170,person211,winter\_0102)}.

kLog learns from interpretations. This means that the data is given as a set of interpretations (or logical worlds) where each interpretation is a set of ground atoms which are true in that world. In this example there are five interpretations: \textsf{ai}, \textsf{graphics}, \textsf{language}, \textsf{systems}, and \textsf{theory}, corresponding to different research groups in the department. 
For instance, a fragment of the interpretation \textsf{ai} is shown in Listing~\ref{lis:uwcse-data}.

\begin{listing}
\begin{Verbatim}[obeytabs=true,fontsize=\footnotesize,fontfamily=helvetica,frame=single,numbers=left,numbersep=3pt,numberblanklines=false,commandchars=\\\{\}]
advised_by(person21,person211).		advised_by(person45,person211).
has_position(person211,faculty).		has_position(person407,faculty).
in_phase(person14,post_generals).	in_phase(person21,post_generals).
in_phase(person284,post_quals).		in_phase(person45,post_generals).
professor(person211).				professor(person407).
publication(title110,person14).		publication(title110,person407).
publication(title25,person21).			publication(title25,person211).
publication(title25,person284).		publication(title25,person407).
publication(title44,person211).		publication(title44,person415).
publication(title44,person45).
student(person14).					student(person21).
student(person284).				student(person45).
ta(course12,person21,winter_0203).	ta(course24,person21,spring_0203).
ta(course24,person70,autumn_0304).
taught_by(course12,person211,autumn_0001).
taught_by(course143,person211,winter_0001).
taught_by(course170,person211,winter_0102).
taught_by(course24,person211,autumn_0102).
taught_by(course24,person211,spring_0203).
taught_by(course24,person407,spring_0304).
taught_by(course82,person407,winter_0102).
\end{Verbatim}
\caption{\label{lis:uwcse-data}
  Database $\mathcal{D}$ for the UW-CSE domain}
\end{listing}
All ground atoms in a particular interpretation together form an instance of a relational database and the overall data set consists of several (disjoint) databases. 

\subsection{Data modeling and learning}
\begin{figure}
  \centering
  \ifarxiv
  \includegraphics[width=.8\textwidth]{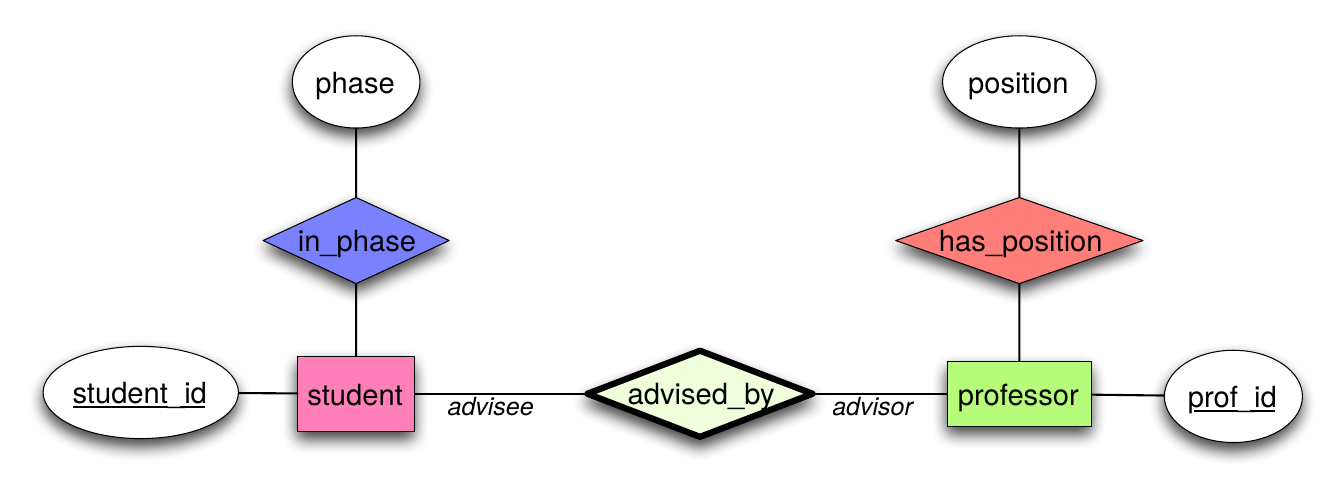}
  \else
  \includegraphics[width=.8\textwidth]{Figures/UW-CSE-Domain}
  \fi
  \caption{\label{fig:UW-CSE-Domain} E/R diagram for the UW-CSE domain.}
\end{figure}

The first step in kLog modeling is to describe the domain using a classic database tool: entity relationship diagrams. We begin by modeling two entity sets: \textsf{student} and \textsf{professor}, two unary relations: \textsf{in\_phase} and \textsf{has\_position}, and one binary relation: \textsf{advised\_by} (which is the target in this example). The diagram is shown in Figure~\ref{fig:UW-CSE-Domain}.
The kLog data model is written using the fragment of code of Listing~\ref{lis:signatures}.
\begin{listing}
\begin{Verbatim}[obeytabs=true,fontsize=\footnotesize,fontfamily=helvetica,frame=single,numbers=left,numbersep=3pt,numberblanklines=false,commandchars=\\\{\}]
  \textit{signature} \textbf{has_position}(professor_id::professor, position::\textit{property})::\textit{extensional}.
  \textit{signature} \textbf{advised_by}(p1::student, p2::professor)::\textit{extensional}.
  \textit{signature} \textbf{student}(student_id::self)::\textit{extensional}.
  \textit{signature} \textbf{professor}(professor_id::self)::\textit{extensional}.
\end{Verbatim}
\caption{\label{lis:signatures}
  Examples of signature declaration.}
\end{listing}

Every entity or relationship that kLog will later use to generate features (see feature generation below) is declared by using the special keyword \textsf{signature}. 
Signatures are similar to the declarative bias used in inductive logic programming systems.
There are two kinds of signatures, annotated by the reserved words \textsf{extensional} and \textsf{intensional}. In the extensional case, all ground atoms have to be listed explicitly in the data file; in the intensional case, ground atoms are defined implicitly using Prolog definite clauses. A signature has a name and a list of arguments with types. A type is either the name of an entity set (declared in some other signature) or the special type \textsf{property} used for numerical or categorical attributes. In the ground atoms, constants that are not properties are regarded as identifiers and are simply used to connect ground atoms (these constants disappear in the result of the graphicalization procedure explained below).\footnote{ The special type name \textsf{self} is used to denote the name of the signature being declared, as in lines 7--8 of Listing \ref{lis:signatures}. Thus, a signature containing an argument of type self introduces a new entity set while other signatures introduce 
relationships.}

One of the powerful features of kLog is its ability to introduce novel relations using a mechanism resembling deductive databases. Such relations are typically a means of injecting domain knowledge. In our example, it may be argued that the likelihood that a professor advises a student increases if the two persons have been engaged in some form of collaboration, such as co-authoring a paper, or working together in teaching activities. 
In Listing~\ref{lis:intensional} we show two intensional signatures for this purpose. An intensional signature declaration must be complemented by a predicate (written in Prolog) which defines the new relation.
When working on a given interpretation, kLog asserts all the true ground atoms in that interpretation in the Prolog database and collects all true groundings for the predicate associated with the intensional signature.
\begin{listing}
\begin{Verbatim}[fontsize=\footnotesize,fontfamily=helvetica,frame=single,numbers=left,numbersep=3pt,numberblanklines=false,commandchars=\\\{\}]
\textit{signature} \textbf{on_same_course}(s::student,p::professor)::\textit{intensional}.
on_same_course(S,P) :-
    professor(P), student(S),
    ta(Course,S,Term), taught_by(Course,P,Term).
 
\textit{signature} \textbf{on_same_paper}(s::student,p::professor)::\textit{intensional}.
on_same_paper(S,P) :-
    student(S), professor(P),
    publication(Pub, S), publication(Pub,P).

\textit{signature} \textbf{n_common_papers}(s::student,p::professor,n::\textit{property})::\textit{intensional}.
n_common_papers(S,P,N) :-
    student(S), professor(P),
    setof(Pub, (publication(Pub, S), publication(Pub,P)), CommonPapers),
    length(CommonPapers,N).
\end{Verbatim}
\caption{\label{lis:intensional}
  Intensional signatures and associated Prolog predicates.}
\end{listing}
Intensional signatures can also be effectively exploited to introduce aggregated attributes~\cite{De-Raedt:2008:Logical-and-relational-learning}. The last signature in Listing~\ref{lis:intensional} shows for example how to count the number of papers a professor and a student have published together.
\subsection{Graphicalization}

\begin{figure}
  \centering
  \ifarxiv
  \includegraphics[width=.9\textwidth]{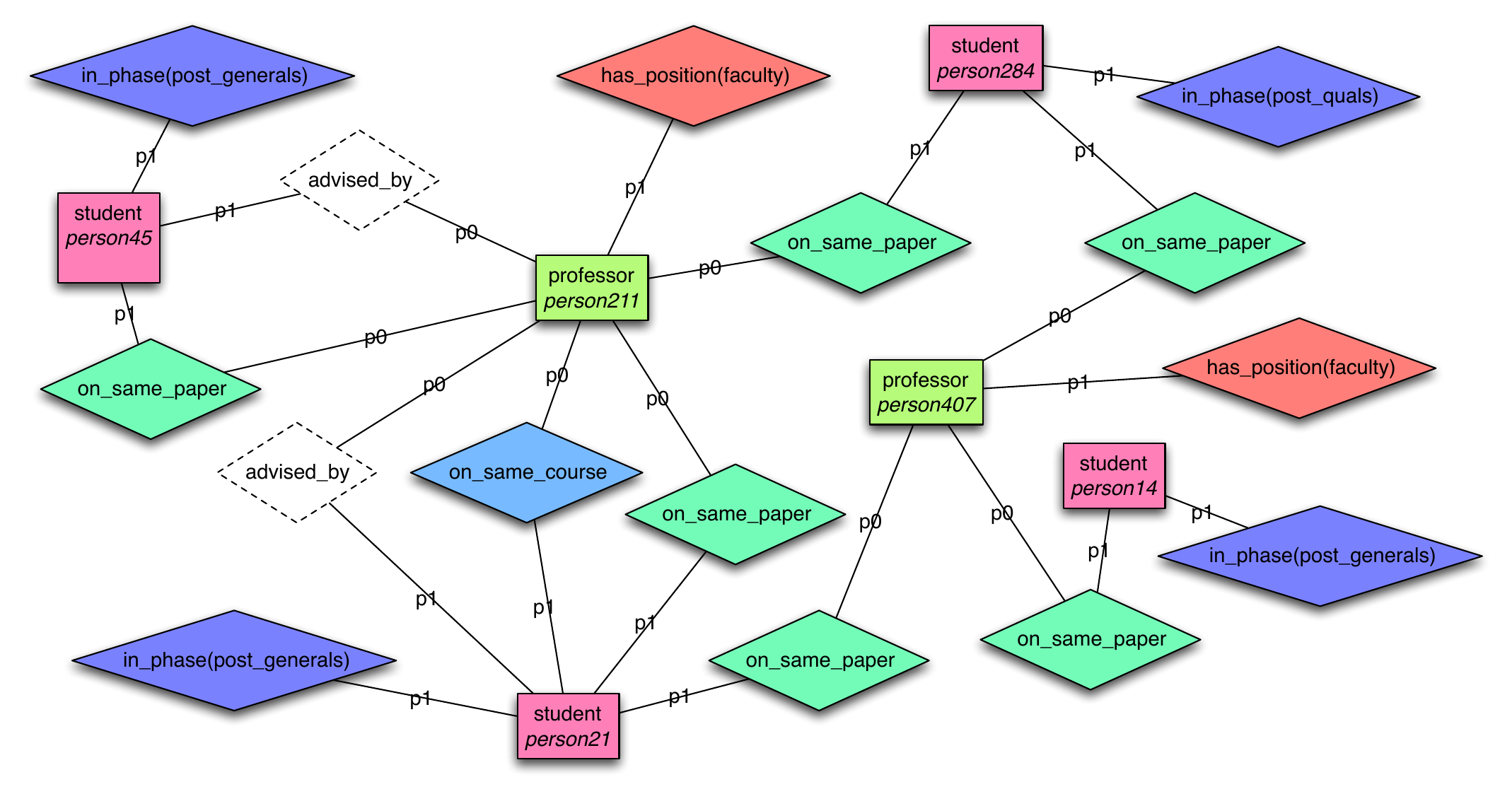}
  \else
  \includegraphics[width=.9\textwidth]{Figures/graphicalization}
  \fi
  \caption{\label{fig:graphicalization} Graphicalization on the tiny data set in the running example. Identifiers (such as person45) are only shown for the sake of clarity; kLog does not use them to generate features.}
\end{figure}

Graphicalization is our approach to capture the relational structure
of the data by means of a graph. It enables the use of kernel methods
in SRL via a graph kernel, which measures the similarity between two
graphs.
The procedure maps a set of ground atoms into a bipartite undirected graph whose nodes are true ground atoms and whose edges connect an entity atom to a relationship atom if the identifier of the former appears as an argument in the latter.
The graph resulting from the graphicalization of the \textsf{ai} interpretation is shown in Figure~\ref{fig:graphicalization}\footnote{Note that interpretation predicates which exist in the data but have not a corresponding signature (e.g., \textsf{publication}) do not produce nodes in the graph. However these predicates may be conveniently exploited in the bodies of the intensional signatures (e.g., \textsf{on\_same\_paper} refers to \textsf{publication}).}.
It is from this graph that kLog will generate propositional features (based on a graph kernel) for use in 
the learning procedure.  The details of the graphicalization procedure and the kernel are given in Sections~\ref{sec:graphicalization} and~\ref{sec:kernel}, respectively.

Now that we have specified the inputs to the learning system, we still
need to determine the learning problem.  This is declared in kLog by
designating one (or more) signature(s) as target (in this domain, the
target relation is \textsf{advised\_by}). Several library predicates
are designed for training, e.g., \textsf{kfold} performs a k-fold cross
validation. These predicates accept a list of target signatures which
specifies the learning problem.

\section{Statistical setting}
\label{sec:statistical-modeling}

Our general approach to construct a statistical model 
is based on the learning from interpretations
setting~\cite{De-Raedt:2008:Logical-and-relational-learning}. An
interpretation, or logical world, is a set of ground atoms $z$.
We assume that interpretations are sampled identically and
independently from a fixed and unknown distribution $D$. We denote by
$\{z_i; i\in\mathcal{I}\}$ the resulting data set, where $\mathcal{I}$
is a given index set (e.g., the first $n$ natural numbers) that can be
thought of as interpretation identifiers. Like in other statistical
learning systems, the goal is to use a data sample to make some form
of (direct or indirect) inference about this distribution.  

For the sake of simplicity, throughout this paper we will mainly focus
on supervised learning.
In this case, it is customary to think of data
as consisting of two separate portions: \textit{inputs} (called
predictors or independent variables in statistics) and
\textit{outputs} (called responses or dependent variables). In our
framework, this distinction is reflected in the set of ground atoms in
a given interpretation.  That is, $z$ is partitioned into two sets:
$x$ (input or evidence atoms) and $y$ (output or query atoms).

The goal of supervised learning in this setting is to construct a
prediction function $f$ that maps the set of input atoms $x$ (a
partial interpretation) into a set of output atoms $f(x)$.  To this
end, a feature vector\footnote{Our approach for constructing feature
  vectors is explained in Section~\ref{sec:graphicalization}.}
$\phi(x,y)$ is associated with each complete interpretation $z=(x,y)$.
A \textit{potential} function based on the linear model
$F(x,y)=w\transpose\phi(x,y)$ is then used to ``score'' the
interpretation. Prediction (or inference) is the process of maximizing
$F$ with respect to $y$, i.e. $f(x)=\argmax_y F(x,y)$.  Learning is
the process of fitting $w$ to the available data, typically using some
statistically motivated loss function that measures the discrepancy
between the prediction $f(x_i)$ and the observed output $y_i$ on the
$i$-th training
interpretation. 
This setting is related to other kernel-based approaches to structured
output learning
(e.g.,~\cite{Tsochantaridis:2006:Large-margin-methods}), which may be
developed without associating a probabilistic interpretation to $F$.

The above perspective covers a number of commonly used algorithms
ranging from propositional to relational learning. To see this,
consider first binary classification of categorical attribute-value
data. In this case, models such as naive Bayes, logistic regression,
and support vector machines (SVM) can all be constructed to share
exactly the same feature space. Using indicator functions on attribute
values as features,
the three models use a hyperplane as their decision function:
$f(x)=\argmax_{y\in\{\mathsf{false},\mathsf{true}\}}
w\transpose\phi(x,y)$ 
where for naive Bayes, the joint probability of $(x,y)$ is
proportional to $\exp(w\transpose\phi(x,y))$.
The only
difference between the three models is actually in the way $w$ is
fitted to data: SVM optimizes a regularized functional based on the
hinge loss function, logistic regression maximizes the conditional
likelihood of outputs given inputs (which can be seen as minimizing a
smoothed version of the SVM hinge loss), and naive Bayes maximises the
joint likelihood of inputs and outputs. The last two models are often
cited as an example of generative-discriminative conjugate pairs
because of the above
reasons~\cite{Ng:2002:On-discriminative-vs.-generative-classifiers:}. 

When moving up to a slightly richer data type like sequences (perhaps
the simplest case of relational data), the three models have well
known extensions: naive Bayes extends to hidden Markov models (HMMs),
logistic regression extends to conditional random fields
(CRFs)~\cite{Sutton:2007:An-Introduction-to-Conditional-Random}, and
SVM extends to structured output SVM for
sequences~\cite{Altun:2003:Hidden-markov-support,Tsochantaridis:2006:Large-margin-methods}.
Note that when HMMs are used in the supervised learning setting (in
applications such as part-of-speech tagging) the observation $x$ is
the input sequence, and the states form the output sequence $y$ (which
is observed in training data). In the simplest version of these three
models, $\phi(x,y)$ contains a feature for every pair of states
(transitions) and a feature for every state-observation pair
(emissions). Again, these models all use the same feature space
(see, e.g.,~\cite{Sutton:2007:An-Introduction-to-Conditional-Random} for
a detailed discussion).  

When moving up to arbitrary relations, the three above models can
again be generalized but many complications arise, and the large
number of alternative models suggested in the literature has
originated the SRL alphabet soup mentioned at the beginning of the paper.  Among generative
models, one natural extension of HMMs is stochastic context free
grammars~\cite{Lari:1991:Applications-of-stochastic-context-free},
which in turn can be extended to 
stochastic logic
programs~\cite{Muggleton:1996:Stochastic-logic-programs}. More
expressive systems include probabilistic relational models
(PRMs)~\cite{friedman99:learn} and Markov logic networks
(MLNs)~\cite{Richardson:2006:Markov-logic-networks}, when trained
generatively.  Generalizations of SVM for relational structures akin
to context free grammars have also been 
investigated~\cite{Tsochantaridis:2006:Large-margin-methods}. Among
discriminative models, CRFs can be extended from linear chains to
arbitrary
relations~\cite{Sutton:2007:An-Introduction-to-Conditional-Random},
for example in the form of discriminative Markov
networks~\cite{Taskar:2002:Discriminative-probabilistic-models} and
discriminative Markov logic
networks~\cite{Richardson:2006:Markov-logic-networks}. The use of
SVM-like loss functions has also been explored in max-margin Markov
networks (M$^3$N)~\cite{Taskar:2003:Max-margin-Markov-networks}.
These models can cope with relational data by adopting a richer
feature space.  kLog contributes to this perspective as it is a language 
for generating a set of features starting from a logical and
relational learning problem and using these features for learning a
(linear) statistical model. 
\begin{table}
  \centering
  \caption{Relationship among some statistical relational learning models.}
  \label{tab:trilogy}
\begin{tabular}{lll}
Propositional & Sequences & General relations \\
\hline
Naive Bayes & HMM & Generative MLN \\
Logistic regression & CRF & Discriminative MLN \\
SVM & SVM-HMM & M$^3$N  \\
\end{tabular}
\end{table}

Table~\ref{tab:trilogy} shows the relationships among some of these
approaches. Methods on the same row use similar loss functions, while
methods in the same column can be arranged to share the same feature
space. In principle, kLog features can be used with any loss.

\section{The kLog language}
\label{sec:klog-language}
A kLog program consists of:
\begin{itemize}
\item a set of ground facts, embedded in a standard
  Prolog database, representing the \textbf{data} of the learning
  problem (see, e.g., Listing~\ref{lis:uwcse-data});
\item a set of signature declarations (e.g., Listing~\ref{lis:signatures});
\item a set of Prolog predicates associated with intensional signatures (e.g., Listing~\ref{lis:intensional});
\item a standard Prolog program which specifies the learning problem
  and makes calls to kLog library predicates.
\end{itemize}

\subsection{Data modeling and signatures}
\label{sec:data_modeling}
In order to specify the semantics of the language, it is convenient to
formalize the domain of the learning problem as a set of constants
(objects) $\mathcal{C}$ and a finite set of relations $\mathcal{R}$.
Constants are partitioned into a set of \textit{entity identifiers}
$\mathcal{E}$ (or identifiers for short) and set of \textit{property
  values} $\mathcal{V}$.  Identifiers are themselves partitioned into
$k$ \textit{entity-sets} $\mathcal{E}_1,\dots,\mathcal{E}_k$.
A ground atom $r(c_1,\dots,c_n)$ is a relation symbol (or predicate
name) $r$ of arity $n$ followed by an $n$-tuple of constant symbols
$c_i$. An interpretation (for the learning problem) is a
\textit{finite} set of ground atoms.

The \textit{signature} for a relation $r/m\in\mathcal{R}$ is an expression of the form
$$
r (\mathsf{name}_1::\mathsf{type}_1, \dots,
\mathsf{name}_m::\mathsf{type}_m) :: \mathsf{level}
$$ 
where, for all $j=1,\dots,m$,
$\mathsf{type}_j\in\{\mathcal{E}_1,\dots,\mathcal{E}_k,\mathcal{V}\}$
and $\mathsf{name}_j$ is the name of the $j$-th column of $r$. 
If column $j$ does not have type $\mathcal{V}$, then its name can
optionally include a \textit{role field} using the syntax
$\mathsf{name}_j\mathsf{@}\mathsf{role}_j$. If unspecified,
$\mathsf{role}_j$ is set to $j$ by default. The level of a signature
is either \textsf{intensional} or \textsf{extensional}. In the
extensional case, all the atoms which contribute to every
interpretation are those and only those listed in the data.
In the intensional case, ground atoms are those which
result from the Prolog intensional predicates using Prolog semantics
(optionally with tabling). In this way, users
may take advantage of many of the extensions of definite
clause logic that are built into Prolog.

The ability to specify intensional predicates through clauses 
(see an example in Listing~\ref{lis:intensional}) is most
useful for introducing background knowledge in the learning
process and common practice in inductive logic programming \cite{De-Raedt:2008:Logical-and-relational-learning}.  As explained in Section~\ref{sec:kernel}, features for the
learning process are derived from a graph whose vertices are ground
facts in the database; hence the ability to declare rules that
specify relations directly translates into the ability to design
and maintain features in a declarative fashion. This is a key
characteristic of kLog and, in our opinion, one of the key reasons
behind the success of related systems like Markov logic.

In order to ensure the well-definedness of the subsequent
graphicalization procedure (see Section~\ref{sec:graphicalization}),
we introduce two additional database assumptions. First, we
require that the primary key of every relation consists of the columns
whose type belongs to $\{\mathcal{E}_1,\dots,\mathcal{E}_k\}$
(i.e., purely of identifiers).  This is perhaps the main difference
between the kLog and the E/R data models.  As it will become more
clear in the following, identifiers are just placeholders and are kept
separate from property values so that learning algorithms will not
rely directly on their values to build a decision function\footnote{To
  make an extreme case illustrating the concept, it makes no sense to
  predict whether a patient suffers from a certain disease based on
  his/her social security number.}.

The \textit{relational arity} of a relation is the length of its
primary key. As a special case, relations of zero relational arity are
admitted and they must consist of at most a single ground atom in any
interpretation\footnote{This kind of relations is useful to represent
  \textit{global} properties of an interpretation (see
  Example~\ref{ex:iid}).}.  Second, for every entity-set
$\mathcal{E}_i$ there must be a distinguished relation $r/m
\in\mathcal{R}$ that has relational arity 1 and key of type
$\mathcal{E}_i$. These distinguished relations are called
\textit{E-relations} and are used to introduce entity-sets, possibly
with $(m-1)$ attached properties as satellite data. The remaining
$|\mathcal{R}|-k$ relations are called \textit{R-relations} or
\textit{relationships}\footnote{Note that the word ``relationship''
  specifically refers to the association among entities while
  ``relation'' refers the more general association among entities and
  properties.}, which may also have properties. Thus, primary keys for
R-relations are tuples of foreign keys.

\subsection{Supervised learning jobs}
A supervised learning job in kLog is specified as a set of relations. We begin
defining the semantics of a job consisting of a single
relation. Without loss of generality, let us assume that this relation
has signature
\begin{equation}
\label{eq:jobsignature}
r(\mathsf{name}_1\!::\!\mathcal{E}_{i(1)},\dots,\mathsf{name}_n\!::\!\mathcal{E}_{i(n)},
\mathsf{name}_{n+1}\!::\!\mathcal{V},\dots,\mathsf{name}_{n+m}\!::\!\mathcal{V})
\end{equation}
with $i(j)\in\{1,\dots,k\}$ for $j=1,\dots,n$. Conventionally, if
$n=0$ there are no identifiers and if $m=0$ there are no properties.
Recall that under our assumptions the primary key of $r$ must consist of entity identifiers (the first $n$ columns). Hence, $n>0$ and $m>0$
implies that $r$ represents a function
with domain
$\mathcal{E}_{i(1)} \times \dots \times \mathcal{E}_{i(n)}$ and range
$\mathcal{V}^{m}$. If $m=0$ then $r$ can be seen as a function with
a Boolean range.

Having specified a target relation $r$, kLog is able to infer the
partition $x\cup y$ of ground atoms into inputs and outputs in the
supervised learning setting. The output $y$ consists of all ground
atoms of $r$ and all ground atoms of any intensional relation $r'$
which depends on $r$. The partition is inferred by analyzing the
dependency graphs of Prolog predicates defining intensional relations,
using an algorithm reminiscent of the call graph computation in
ViPReSS \cite{refactoring:tplp}.

We assume that the training data is a set of complete
interpretations\footnote{kLog could be extended to deal with missing
  data by removing the closed world assumption and requiring some
  specification of the actual false groundings.}.
During prediction, we are given a partial interpretation consisting of
ground atoms $x$, and are required to \textit{complete} the
interpretation by predicting output ground atoms $y$. For the purpose
of prediction accuracy estimation, we will be only interested in the
ground atoms of the target relation (a subset of $y$).
 
Several situations may arise depending on the relational arity $n$ and
the number of properties $m$ in the target relation $r$, as summarized
in Table~\ref{tab:jobs}. When $n=0$, the declared job consists of
predicting one or more properties of an entire interpretation, when
$n=1$ one or more properties of certain entities, when $n=2$ one or
more properties of pairs of entities, and so on. When $m=0$ (no
properties) we have a binary classification task (where positive cases
are ground atoms that belong to the complete
interpretation). Multiclass classification can be properly declared by
using $m=1$ with a categorical property, which ensures mutual
exclusiveness of classes. Regression is also declared by using $m=1$ but
in this case the property should be numeric. Note that property types
(numerical vs. categorical) are automatically inferred by kLog by
inspecting the given training data.
An interesting scenario occurs when
$m>1$ so that two or more properties are to be predicted at the same
time. A similar situation occurs when the learning job consists of
several target relations. kLog recognizes that such a declaration
defines a multitask learning job. However having recognized a
multitask job does not necessarily mean that kLog will have to use a
multitask learning algorithm capable of taking advantage of
correlations between tasks (like, e.g.,~\cite{argyriou2008convex}). This
is because, by design and in line with the principles of declarative
languages, kLog separates ``what'' a learning job looks like and
``how'' it is solved by applying a particular learning algorithm. We
believe that the separation of concerns at this level permits greater
flexibility and extendability and facilitates plugging-in alternative
learning algorithms (a.k.a.\ kLog models) that have the ability to
provide a solution to a given job.

\begin{table}
  \caption{\label{tab:jobs}Single relation jobs in kLog.}
  \centering
  \small
\begin{tabular}{p{0.13\textwidth}p{0.24\textwidth}p{0.24\textwidth}p{0.24\textwidth}r}
 & \multicolumn{3}{c}{Relational arity, $n$}\\
\# of properties, $m$ & \centering 0 & \centering 1 & \centering 2 & ~ \\
\hline
\centering 0 & Binary classification of interpretations & Binary classification of entities & Link prediction &  \\
\centering 1 & Multiclass / regression on interpretations & Multiclass / regression on entities & Attributed link prediction &  \\
\centering $>$1 & Multitask on interpretations & Multitask predictions on entities & Multitask attrib\-uted link prediction &  \\
\end{tabular}
\end{table}

\subsection{Implementation}
kLog is currently embedded in Yap Prolog~\cite{costa2012yap} and
consists of three main components: (1) a domain-specific interpreter,
(2) a database loader, and (3) a library of predicates that are used
to specify the learning task, to declare the graph kernel and the
learning model, and to perform training, prediction, and performance
evaluation. The domain-specific interpreter parses the signature
declarations (see Section~\ref{sec:data_modeling}), possibly enriched
with intensional and auxiliary predicates. The database loader reads
in a file containing extensional ground facts and generates a graph
for each interpretation, according to the procedure detailed in
Section~\ref{sec:graphicalization}. The library includes common
utilities for training and testing. Most of kLog is written in Prolog
except feature vector generation and the statistical learners, which
are written in C++. kLog incorporates LibSVM~\cite{libsvm:other} and Stochastic
gradient descent~\cite{Bottou2010:proc} and can interface with arbitrary 
(external) SVM solvers by coding appropriate data conversion wrappers.

\section{Examples}
\label{sec:examples}
In Section~\ref{sec:running-example} we have given a detailed example of kLog in a link prediction problem.
The kLog setting encompasses a larger ensemble of machine
learning scenarios, as detailed in the following examples, which we
order according to growing complexity of the underlying learning task.

\begin{example}[Classification of independent interpretations]
  \label{ex:iid}
  This is the simplest supervised learning problem with structured
  input data and scalar (unstructured) output. For the sake of
  concreteness, let us consider the problem of small molecule
  classification as pioneered in the relational learning setting in
  ~\cite{Srinivasan:1994:Mutagenesis:-ILP-experiments-in-a-non-determinate}. This
  domain is naturally modeled in kLog as follows. Each molecule
  corresponds to one interpretation; there is one E-relation,
  \textnormal{\textsf{atom}}, that may include properties such as
  \textnormal{\textsf{element}} and \textnormal{\textsf{charge}};
  there is one relationship of relational arity 2,
  \textnormal{\textsf{bond}}, that may include a
  \textnormal{\textsf{bond\_type}} property to distinguish among
  single, double, and resonant chemical bonds; there is finally a
  zero-arity relationship, \textnormal{\textsf{active}},
  distinguishing between positive and negative interpretations. A
  concrete example is given in Section~\ref{sec:molecules}.
\end{example}

\begin{example}[Regression and multitask learning]
  \label{ex:regression}
  The above example about small molecules can be extended to the case
  of regression where the task is to predict a real-valued property
  associated with a molecule, such as its biological activity or its
  octanol/water partition coefficient
  (logP)~\cite{Wang:1997:A-new-atom-additive-method-for-calculating}.
  Many problems in quantitative structure-activity relationships
  (QSAR) are actually formulated in this way. The case of regression
  can be handled simply by introducing a target relation with
  signature \textnormal{\textsf{activity(act::property)}}.

  If there are two or more properties to be predicted, one possibility
  is to declare several target relations, e.g., we might add
  \textnormal{\textsf{logP(logp::property)}}. Alternatively we
  may introduce a target relation such as:

  \begin{center}
    \textnormal{\textsf{target\_properties(activity::property,logp::property)}}.
  \end{center}

  Multitask learning can be handled trivially by learning independent
  predictors; alternatively, more sophisticated algorithms that take
  into account correlations amongst tasks (such
  as~\cite{Evgeniou:2006:Learning-multiple-tasks}) could be used.
\end{example}

\begin{example}[Entity classification]
  \label{ex:entity-classification}
  A more complex scenario is the collective classification of several
  entities within a given interpretation.  We illustrate this case
  using the classic WebKB
  domain~\cite{Craven:1998:Learning-to-Extract-Symbolic}. The data set
  consists of Web pages from four Computer Science departments and
  thus there are four interpretations: \textnormal{\textsf{cornell}},
  \textnormal{\textsf{texas}}, \textnormal{\textsf{washington}}, and
  \textnormal{\textsf{wisconsin}}.  In this domain there are two
  E-relations: \textnormal{\textsf{page}} (for webpages) and
  \textnormal{\textsf{link}} (for hypertextual links).  Text in each
  page is represented as a bag-of-words (using the R-relation
  \textnormal{\textsf{has}}) and hyperlinks are modeled by the
  R-relations \textnormal{\textsf{link\_to}} and
  \textnormal{\textsf{link\_from}}.
  Text associated with hyperlink anchors is represented by the
  R-relation \textnormal{\textsf{has\_anchor}}.

  The goal is to classify each Web page.  There are different data
  modeling alternatives for setting up this classification task. One
  possibility is to introduce several unary R-relations associated
  with the different classes, such as \textnormal{\textsf{course}},
  \textnormal{\textsf{faculty}}, \textnormal{\textsf{project}}, and
  \textnormal{\textsf{student}}. The second possibility is to add a
  property to the entity-set \textnormal{\textsf{page}}, called
  \textnormal{\textsf{category}}, and taking values on the different
  possible categories. It may seem that in the latter case we are just
  reifying the R-relations describing categories. However there is an
  additional subtle but important difference: in the first modeling
  approach it is perfectly legal to have an interpretation where a
  page belongs simultaneously to different categories. This becomes
  illegal in the second approach since otherwise there would be two or
  more atoms of the E-relation \textnormal{\textsf{page}} with the
  same identifier.

  From a statistical point of view, since pages for the same
  department are part of the same interpretation and connected by
  hyperlinks, the corresponding category labels are interdependent
  random variables and we formally have an instance of a supervised
  structured output problem~\cite{BakIr:2007:Predicting-stru},
  that in this case might  also be referred to as \textit{collective
    classification}~\cite{Taskar:2002:Discriminative-probabilistic-models}. There
  are however studies in the literature that consider pages to be
  independent
  (e.g.,~\cite{Joachims:2002:Learning-to-classify-text}).
\end{example}

\begin{example}[One-interpretation domains]
\label{ex:one-interpretation}
We illustrate this case on the Internet Movie Database (IMDb) data
set. Following the setup
in~\cite{Neville:2003:Collective-classification-with}, the problem is
to predict ``blockbuster'' movies, i.e., movies that will earn more
than \$2 million in their opening weekend. The entity-sets in this domain
are \textnormal{\textsf{movie}}, 
\textnormal{\textsf{studio}}, and
\textnormal{\textsf{individual}}.
Relationships include
\textnormal{\textsf{acted\_in(actor::individual, m::movie)}},
\textnormal{\textsf{produced(s::studio, m::movie)}}, and
\textnormal{\textsf{directed(director::individual, m::movie)}}. The target unary
relation \textnormal{\textsf{blockbuster(m::movie)}} collects positive
cases. Training in this case uses a partial interpretation (typically
movies produced before a given year). When predicting the class of
future movies, data about past movies' receipts can be used to
construct features (indeed, the count of blockbuster movies produced
by the same studio is one of the most informative
features~\cite{Frasconi:2008:Feature-discovery-with}).

A similar scenario occurs for protein function prediction. Assuming
data for just a single organism is available, there
is one entity set (\textnormal{\textsf{protein}}) and a binary
relation \textnormal{\textsf{interact}} expressing protein-protein
interaction~\cite{Vazquez:2003:Global-protein-function,Lanckriet:2004:Kernel-based-data-fusion}.
\end{example}


\section{Graphicalization and feature generation}
\label{sec:graphicalization}
The goal is to map an interpretation $z=(x,y)$ into a feature vector
$\phi(z)=\phi(x,y)\in\mathcal{F}$ (see Section~\ref{sec:statistical-modeling}). This enables the application of
several supervised learning algorithms that construct linear functions
in the feature space $\mathcal{F}$. In this context, $\phi(z)$ can be
either computed explicitly or defined implicitly, via a kernel
function $K(z,z')=\langle \phi(z),\phi(z') \rangle$.  Kernel-based
solutions are very popular, sometimes computationally faster, and
can exploit infinite-dimensional feature spaces. On the other hand, explicit
feature map construction may offer advantages in our setting, in
particular when dealing with large scale learning problems (many
interpretations) and structured output tasks (exponentially many
possible predictions). Our framework is based on two steps: first an
interpretation $z$ is mapped into an undirected labeled graph $G_z$;
then a feature vector $\phi(z)$ is extracted from
$G_z$. Alternatively, a kernel function on pairs of graphs
$K(z,z')=K(G_z,G_{z'})$ could be computed. The corresponding potential
function is then defined directly as $F(z)=w\transpose\phi(z)$ or as a
kernel expansion $F(z)=\sum_ic_iK(z,z_i)$.

The use of an intermediate graphicalized representation is novel in the context of propositionalization, a  well-known technique in logical and relational
learning~\cite{De-Raedt:2008:Logical-and-relational-learning} that
transforms graph-based 
or relational data 
{\em directly} into an attribute-value
learning format, or possibly into a multi-instance learning one\footnote{In multi-instance learning~\cite{Dietterich1997}, the
  examples are sets of attribute-value tuples or sets of feature
  vectors.}. 
This typically results in a loss of information, 
cf.~\cite{De-Raedt:2008:Logical-and-relational-learning}.  
Our approach transforms the relational data into an {\em equivalent} graph-based format,
without loss of information. After graphicalization, kLog uses the results on kernels for graph-based data to derive 
an explicit high-dimensional feature-based representation.  
Thus kLog directly upgrades these graph-based kernels to a fully relational representation.
Furthermore, there is an extensive literature on graph kernels and
virtually all existing solutions can be plugged into the learning from
interpretations setting with minimal effort. This includes
implementation issues but also the ability to reuse existing
theoretical analyses. Finally, it is notationally simpler to describe
a kernel and feature vectors defined on graphs, than to describe the
equivalent counterpart using the Datalog notation.

The graph kernel choice implicitly determines how predicates' attributes
are combined into features. 

\subsection{Graphicalization procedure}
\label{sec:graphicalization-procedure}
Given an interpretation $z$, we construct a bipartite graph
$G_z([V_z,F_z],E_z)$ as follows (see~\ref{sec:definitions} for
notational conventions and Figure~\ref{fig:graphicalization} for an example). 
\begin{description}
\item[Vertices:] there is a vertex in $V_z$ for every ground atom of
  every $E$-relation, and there is a vertex in $F_z$ for every
  ground atom of every $R$-relation. Vertices are labeled by the predicate name
  of the ground atom, followed by the list of property values.
  Identifiers in a ground atom do not appear in the labels but they
  uniquely identify vertices. The tuple $\mathrm{ids}(v)$ denotes the
  identifiers in the ground atom mapped into vertex $v$.
\item[Edges:] $uv \in E_z$ if and only if $u \in V_z$, $v \in F_z$,
  and $\mathrm{ids}(u) \subset \mathrm{ids}(v)$.  The edge $uv$ is
  labeled by the role under which the identifier of $u$ appears in
  $v$ (see Section~\ref{sec:data_modeling}).
\end{description}
Note that, because of our data modeling assumptions (see
Section~\ref{sec:data_modeling}), the degree of every vertex $v\in
F_z$ equals the relational arity of the corresponding $R$-relation.
The degree of vertices in $V_z$ is instead unbounded and may grow really
large in some domains (e.g., social networks or World-Wide-Web
networks).
The graphicalization process can be nicely interpreted as the
\textit{unfolding} of an E/R diagram over the data, i.e., the E/R
diagram is a template that is expanded according to the given
ground atoms (see Figure~\ref{fig:graphicalization}). There are several
other examples in the literature where a graph template is expanded
into a ground graph, including the plate notation in graphical
models~\cite{Koller:2009:Probabilistic-graphical-models:}, encoding
networks in neural networks for learning data
structures~\cite{Frasconi:1998:A-general-framework-for-adaptive}, and
the construction of Markov networks in Markov
logic~\cite{Richardson:2006:Markov-logic-networks}. The semantics of
kLog  graphs for the learning procedure is however quite
different and intimately related to the concept of graph kernels, as
detailed in the following section.

\subsection{Graph kernel}
\label{sec:kernel}
Learning in kLog is performed using a suitable graph kernel on the
graphicalized interpretations. While in principle any graph kernel can
be employed, there are several requirements that the chosen kernel has
to meet in practice. On the one hand, computational efficiency is very
important, both with respect to the number of graphs, and with respect
to the graph size, as the grounding phase in the graphicalization
procedure can yield very large graphs. On the other hand, we need a
general purpose kernel with a flexible bias to adapt to a wide variety
of application scenarios. In the case of independent interpretations
(as in Examples~\ref{ex:iid} and~\ref{ex:regression}), any graph
kernel (several exist in the literature, see
e.g. \cite{horvath04,ralaivola2005graph,mahe2005gkm,Gartner:2008:Kernels-for-str,Vishwanathan2010:jrnl,Costa::Fast-neighborhood-subgraph,shervashidze2011weisfeiler}
and references therein) can be directly applied to the result of the
graphicalization procedure since there is exactly one graph for each
interpretation. However, when the task requires to make predictions
about tuples of entities within the same interpretation (as
in Section~\ref{sec:running-example} or in
Examples~\ref{ex:entity-classification}
and~\ref{ex:one-interpretation}) an immediate application of existing
graph kernels (such as those in the above references) is not
straightforward. Additionally, if the graph has vertices with
large degree (as in Example~\ref{ex:entity-classification}),
kernels based on hard subgraph matching may severely overfit the data.

In the current implementation, we use an extension of
NSPDK~\cite{Costa::Fast-neighborhood-subgraph}.  While the original
kernel is suitable for sparse graphs with discrete vertex and edge
labels, here we propose extensions to deal with soft matches
(Section~\ref{sec:soft-matches}) and a larger class of graphs whose
labels are tuples of mixed discrete and numerical types
(Section~\ref{sec:tuples}). In Section~\ref{sec:viewpoints} we finally
introduce a variant of NSPDK with viewpoints, which can handle the
case of multiple predictions within the same interpretation.  Similar
extensions could be devised for other kinds of kernels but we do not
discuss them in this paper.


\subsubsection{Kernel definition and notation}
\label{sec:nspdk}
The NSPDK is an instance of a decomposition kernel, where {\em ``parts''}
are pairs of subgraphs (for more details on decomposition kernels see
\ref{sec:decomposition_kernels}). For a given graph $G=(V,E)$, and an
integer $r\geq 0$, let $N_r^v(G)$ denote the subgraph of $G$ rooted in
$v$ and induced by the set of vertices
\begin{equation}
\label{eq:neighborhood-v}
V^v_r = \{x\in V:d^\star(x,v)\leq r\}, 
\end{equation}
where $d^\star(x,v)$ is the shortest-path distance between $x$ and $v$
\footnote{Conventionally $d^\star(x,v)=\infty$ if no path exists between $x$
and $v$}. A neighborhood is therefore a topological {\em ball} with
center $v$.  Let us also introduce the following
\textit{neighborhood-pair} relation:
\begin{equation}
  \label{eq:neighborhood-pair}
  R_{r,d} = \{(N_r^v(G),N_r^u(G),G):d^\star(u,v)=d\}
\end{equation}
that is, relation $ R_{r,d}$ identifies pairs of neighborhoods of
radius $r$ whose roots are exactly at distance $d$. We define $\kappa_{r,d}$
over graph pairs as the decomposition kernel on the relation
$R_{r,d}$, that is:
\begin{equation}
\label{eq:kappard}
\kappa_{r,d}(G,G')=\!\!\!\!\!\!\!\!\!\!\!\! \sum_{
 \begin{scriptsize} 
   \begin{array}{c} 
     A,B \in R_{r,d}^{-1}(G)\\
     A',B' \in R_{r,d}^{-1}(G')
 \end{array} 
 \end{scriptsize} }\!\!\!\!\!\!\!\!\!\!\!\!
\kappa((A,B),(A',B'))
\end{equation}
where $R_{r,d}^{-1}(G)$ indicates the multiset of all pairs of
neighborhoods of radius $r$ with roots at distance $d$ that exist in
$G$.

We can now obtain a flexible parametric family of kernel functions by
specializing the kernel $\kappa$. The general structure of $\kappa$
is:
\begin{equation}
\kappa((A,B),(A',B'))= \kappa_{root}((A,B),(A',B')) \kappa_{subgraph}((A,B),(A',B')).
\end{equation}
In the following, we assume 
\begin{equation}
\kappa_{root}((A,B),(A',B'))=\mathbf{1}_{\ell(r(A))=\ell(r(A'))}\mathbf{1}_{\ell(r(B))=\ell(r(B'))}
\end{equation}
where $\mathbf{1}$ denotes the indicator function, $r(A)$ is the root
of $A$ and $\ell(v)$ the label of vertex $v$. The role of
$\kappa_{root}$ is to ensure that only neighborhoods centered on the
same type of vertex will be compared. Assuming a valid kernel for
$\kappa_{subgraph}$ (in the following Sections we give details on
concrete instantiations), we can finally define the NSPDK as:
\begin{equation}
\label{eq:kernel}
K(G,G')=\sum_r\sum_d\kappa_{r,d}(G,G').
\end{equation}
For efficiency reasons we consider the zero-extension of $K$ obtained
by imposing an upper bound on the radius and the distance parameter:
$K_{\maxradius,\maxdistance}(G,G')=\sum_{r=0}^{\maxradius} \sum_{d=0}^{\maxdistance}
\kappa_{r,d}(G,G')$, that is, we limit the sum of the $\kappa_{r,d}$
kernels for all increasing values of the radius (distance) parameter
up to a maximum given value \maxradius\ (\maxdistance).  Furthermore we consider a
normalized version of $\kappa_{r,d}$, that is: $
\hat{\kappa}_{r,d}(G,G')=\frac{\kappa_{r,d}(G,G')}{\sqrt{\kappa_{r,d}(G,G)
    \kappa_{r,d}(G',G')}}, $ to ensure that relations induced by all
values of radii and distances are equally weighted regardless of the
size of the induced part sets.

Finally, it is easy to show that the NSPDK is a valid kernel as: 1) it is built as a
decomposition kernel over the countable space of all pairs of
neighborhood subgraphs of graphs of finite size; 2) the kernel over
parts is a valid kernel; 3) the zero-extension to bounded
values for the radius and distance parameters preserves positive
definiteness and symmetry; and 4) so does the normalization step.

\begin{figure}
  \centering
  \ifarxiv
  \includegraphics[width=0.95\textwidth]{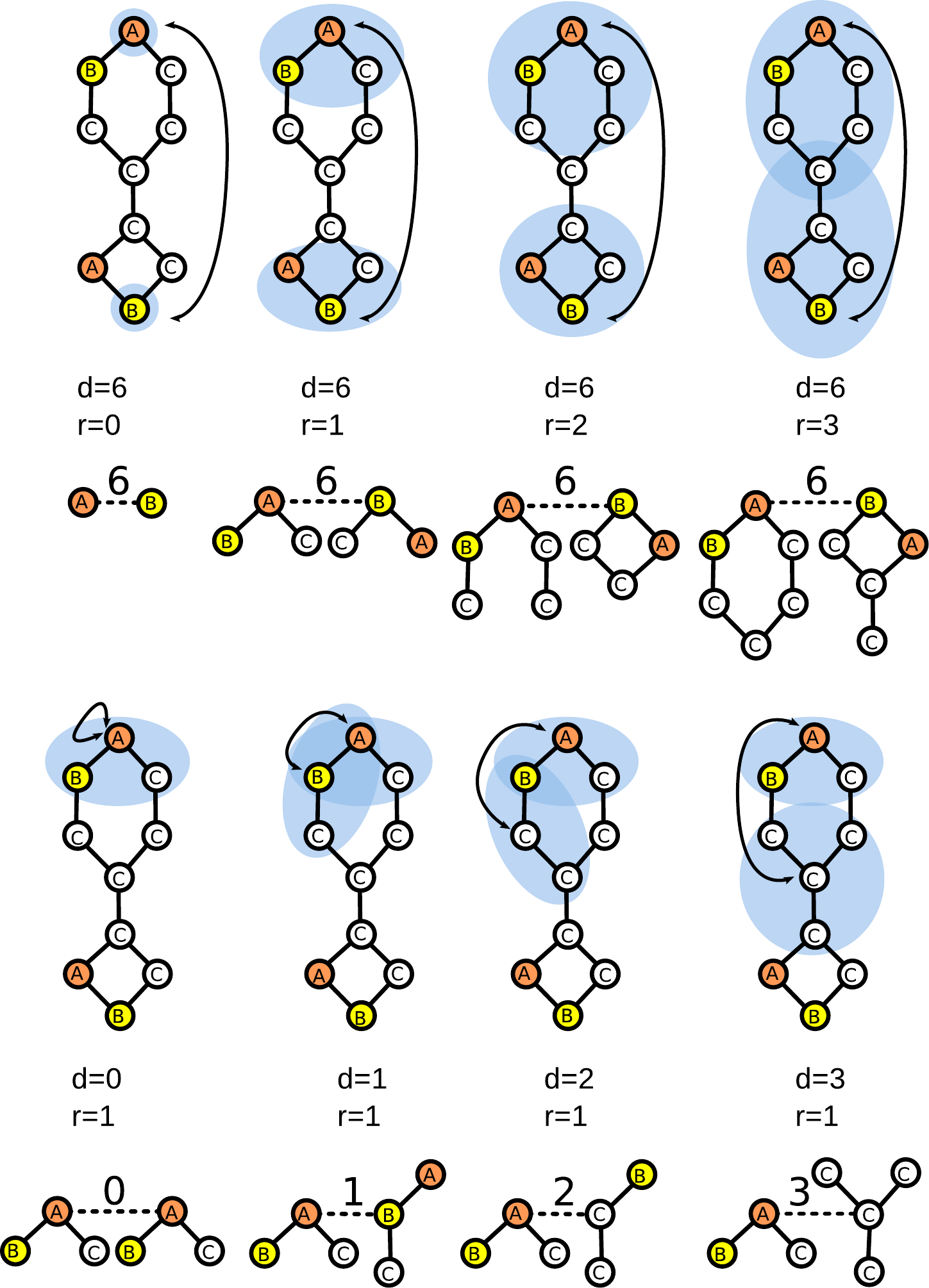}
  \else
  \includegraphics[width=0.95\textwidth]{Figures/nspdk.pdf}
  \fi
  \caption{\label{fig:nspdk_features} Illustration of NSPDK features. Top:
    pairs of neighborhood subgraphs at fixed distance 6 for radii
    0, 1, 2, and 3. Bottom: pairs of neighborhood subgraphs of fixed
    radius 1 at distances 0, 1, 2, and 3.}
\end{figure}

\subsubsection{Subgraph kernels}
The role of $\kappa_{subgraph}$ is to compare pairs of neighborhood
graphs extracted from two graphs. The application of the
graphicalization procedure to diverse relational datasets can
potentially induce graphs with significantly different
characteristics. In some cases (discrete property domains) an
exact matching between neighborhood graphs is appropriate, in other
cases however (continous properties domains) it is more
appropriate to use a {\em soft} notion of matching.

In the following sections, we introduce variants of $\kappa_{subgraph}$
to be used when the atoms in the relational dataset can maximally have
a single discrete or continuous property, or when more general tuples
of properties are allowed.

\subsubsection{Exact graph matching}
\label{sec:exact_match}
An important case is when the atoms, that are mapped by the
graphicalization procedure to the vertex set of the resulting graph,
can maximally have a single discrete property. In this case, an atom
$r(c)$ becomes a vertex $v$, 
whose
label is obtained by concatenation of the signature name and the
attribute value.
In this case, $\kappa_{subgraph}$ has the following form:
\begin{equation}
\label{eq:hardm}
\kappa_{subgraph}((A,B),(A',B')) = \mathbf{1}_{A\cong A'} \mathbf{1}_{B\cong B'}
\end{equation}
where $\mathbf{1}$ denotes the indicator function and $\cong$
isomorphism between graphs.  Note that $\mathbf{1}_{A\cong A'}$ is a
valid kernel between graphs under the feature map $\phi_{\mathrm{cl}}$
that transforms $A$ into $\phi_{\mathrm{cl}}(A)$, a sequence of all
zeros except the $i$-th element equal to 1 in correspondence to the
identifier for the \textit{canonical representation} of
$A$~\cite{McKay81,Yan02}.

Evaluating the kernel in Equation~\ref{eq:hardm} requires as a
subroutine graph isomorphism, a problem for which it is unknown
whether polynomial algorithms exist. Algorithms that are in the
worst case exponential but that are fast in practice do
exist~\cite{McKay81,Yan02}. For special graph classes, such as bounded
degree graphs~\cite{Luks82}, there exist polynomial time algorithms.
However, since it is hard to limit the type of graph produced by the
graphicalization procedure (e.g., cases with very high vertex degree
are possible as in general an entity atom may play a role in an
arbitrary number of relationship atoms), we prefer an approximate
solution with efficiency guarantees based on topological distances
similar in spirit to \cite{sorlin08}.

The key idea is to compute an integer pseudo-identifier for each graph
such that isomorphic graphs are guaranteed to bear the same number
(i.e., the function is a graph invariant), but non-isomorphic graphs are
likely to bear a different number. A trivial identity test between the
pseudo-identifiers then approximates the isomorphism test. 
\ref{sec:graphinvariant} details the computation of the pseudo-identifier.



\subsubsection{Soft matches}
\label{sec:soft-matches}
The idea of counting exact neighborhood subgraphs matches to express
graph similarity is adequate when the graphs are sparse (that is, when
the edge and the vertex set sizes are of the same order) and when the
maximum vertex degree is low. However, when the graph is not sparse or
some vertices exhibit large degrees, the likelihood that two
neighborhoods match exactly quickly approaches zero, yielding a diagonal dominant kernel prone to overfitting\footnote{A concrete example is when text
  information associated to a document is modeled explicitly,
  i.e., when word entities are linked to a document entity: in this
  case the degree corresponds to the document vocabulary size.}. In
these cases a better solution is to relax the all-or-nothing type of
match, allowing subgraphs to match partially or in a {\em soft} way.
Although there exist several graph kernels that are based on this
type of match, they generally suffer from very high computational
costs \cite{Vishwanathan2010:jrnl}. To ensure efficiency, we use
an idea introduced in the Weighted Decomposition Kernel
\cite{Menchetti05:proc}: given a subgraph, we consider only the
multinomial distribution (i.e., the histogram) of the labels,
discarding all structural information. In the soft match kernel, the
comparison between two pairs of neighborhood subgraphs
is replaced by\footnote{Note
  that the pair of neighborhood subgraphs are considered jointly,
  i.e., the label multisets are extracted independently from each
  subgraph in the pair and then combined together.}:
\begin{equation}
\kappa_{subgraph}((A,B),(A',B')) = 
\!\!\!\!\!\!\!\!\!\!\!\! \sum_{
   \begin{array}{c} 
     v \in V(A) \cup V(B)\\
     v' \in V(A') \cup V(B')
 \end{array}
} \!\!\!\!\!\!\!\!\!\!\!\!
\mathbf{1}_{\ell(v)=\ell(v')}
\end{equation}
where $V(A)$ is the set of vertices of $A$. In words, for each pair of
close neighborhoods, we build a histogram counting the vertices with
the same label in either of the neighborhood
subgraphs. The kernel is then computed as the dot product of the
corresponding histograms.

\begin{figure}
  \centering
  \ifarxiv
  \includegraphics[width=.9\textwidth]{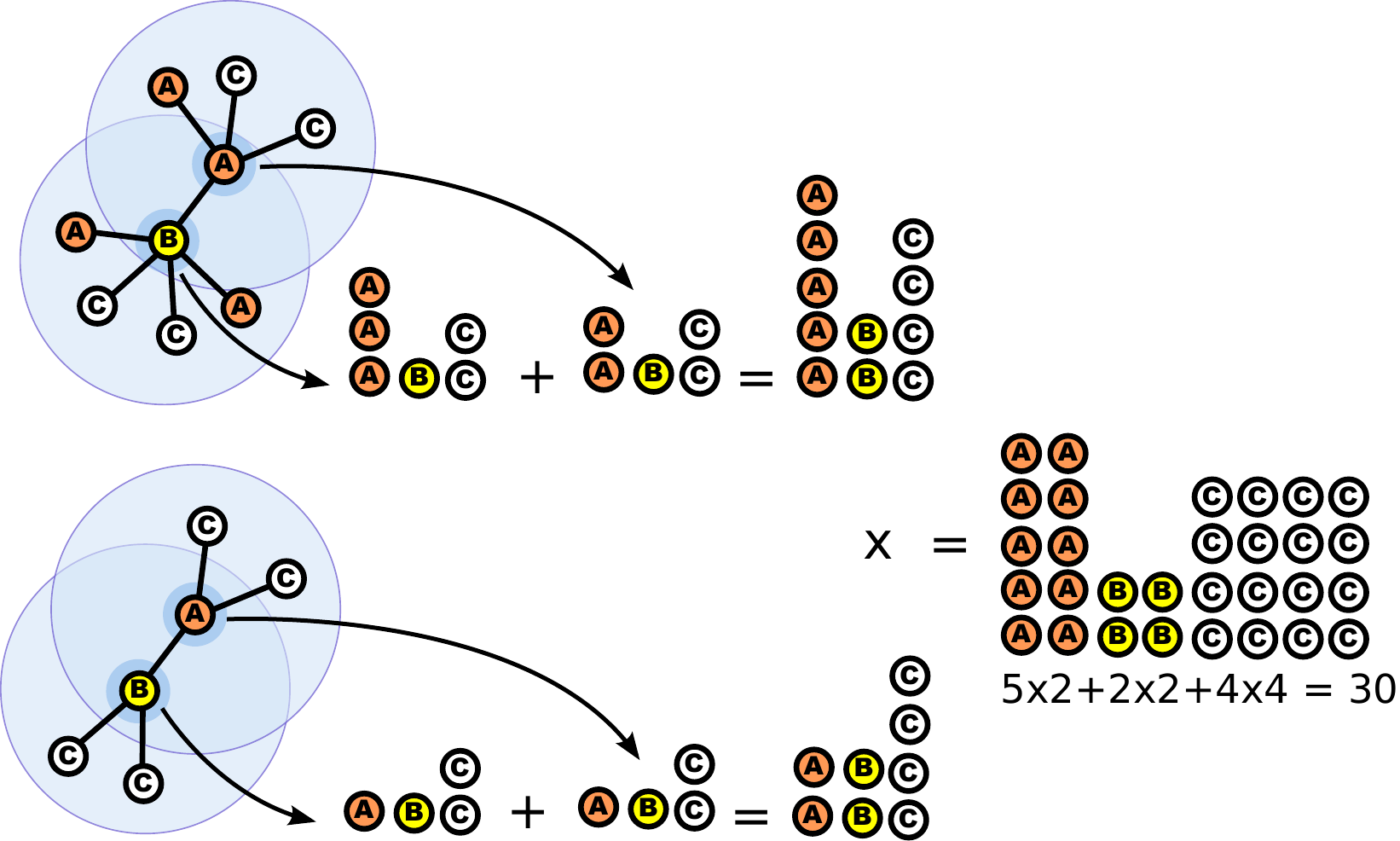}
  \else
  \includegraphics[width=.9\textwidth]{Figures/soft.pdf}
  \fi
  \caption{\label{fig:soft} Illustration of the soft matching
    kernel. Only features generated by a selected pair of vertices are
    represented: vertices A and B at distance 1 yield a multinomial
    distribution of the vertex labels in neighborhoods of radius
    1. On the right we compute the contribution to the kernel value by 
    the represented features.}
\end{figure}

\subsubsection{Tuples of properties}
\label{sec:tuples}
A standard assumption in graph kernels is that vertex and edge labels
are elements of a discrete domain. However, in kLog the information
associated with vertices is a tuple that can contain both discrete and
real values. Here we extend NSPDK to allow both a hard and a
soft match type over graphs with property tuples that can be 
discrete, real, or a mix of both types. 
While similar extensions can be conceived, in principle, for other graph kernels,
the literature has mainly focused on kernels with single categorical labels.
The key idea is to use the
canonical vertex identifier (introduced in Section~\ref{sec:exact_match}) to
further characterize each property: in this way the kernel is defined
only between tuples of vertices that correspond under isomorphism.

The general structure of the kernel on the subgraph can be written as:
\begin{equation}
\label{eq:ksub}
\kappa_{subgraph}((A,B),(A',B')) = 
\!\!\!\!\!\!\!\!\!\!\!\! \sum_{
 \begin{scriptsize} 
   \begin{array}{c} 
     v \in V(A) \cup V(B)\\
     v' \in V(A') \cup V(B')
 \end{array} 
 \end{scriptsize} 
} \!\!\!\!\!\!\!\!\!\!
\mathbf{1}_{\ell(v)=\ell(v')} \kappa_{tuple}(v,v')
\end{equation}
where, for an atom\footnote{We remind the reader that in the
  graphicalization procedure we remove the primary and foreign keys
  from each atom, hence the only information available at the graph
  level are the signature name and the properties values.}
$r(c_1,c_2,\ldots,c_m)$ mapped into vertex $v$, $\ell(v)$ returns the
signature name $r$. $\kappa_{subgraph}$ is a kernel that is defined
over sets of vertices (atoms) and can be decomposed in a part that
ensures matches between atoms with the same signature name, and a
second part that takes into account the tuple of property values.  In
 particular, depending on the type of property values and the type of
 matching required, we obtain the following six cases.

\noindent\textbf{Soft match for discrete tuples.} 
When the tuples contain only discrete elements and one chooses to ignore
the structure in the neighborhood graphs, then each property is treated
independently. Formally:
\begin{equation}
  \kappa_{tuple}(v,v')= \sum_d \mathbf{1}_{prop_d(v)=prop_d(v')}
\end{equation}
where for an atom $r(c_1,c_2,\ldots,c_d,\ldots,c_m)$ mapped into vertex $v$,
$prop_d(v)$ returns the property value $c_d$.

\noindent\textbf{Hard match for discrete tuples.} 
When the tuples contain only discrete elements and one chooses to
consider the structure, then each property key is encoded taking into
account the identity of the vertex in the neighborhood graph and all
properties are required to match jointly. In practice, this is
equivalent to the hard match of
Section~\ref{sec:exact_match} where the property value is replaced
with the concatenation of all property values in the tuple.  Formally,
we replace the label $\ell(v)$ in Equation~\ref{eq:ksub} with the
labeling procedure $\Label^v$ explained in
Section~\ref{sec:exact_match}. In this way, each vertex receives a canonical
label that uniquely identifies it in the neighborhood graphs.  The
double summation of $\mathbf{1}_{\ell(v)=\ell(v')}$ in
Eq.~\ref{eq:ksub} is then performing the selection of the
corresponding vertices $v$ and $v'$ in the two pairs of neighborhood
that are being compared.  Finally, we consider all the elements of the
tuple jointly in order to identify a successful match:
\begin{equation}
  \kappa_{tuple}(v,v')= \prod_d \mathbf{1}_{prop_d(v)=prop_d(v')}.
\end{equation}

\noindent\textbf{Soft match for real tuples.} To upgrade the soft match kernel to
tuples of real values we replace the exact match with the standard
product\footnote{Note that this is equivalent to collecting all
  numerical properties of a vertex's tuple in a vector and then employ
  the standard dot product between vectors.}. The kernel on the tuple
then becomes:
\begin{equation}
\label{eq:soft_real}
  \kappa_{tuple}(v,v')= \sum_c prop_c(v) \cdot prop_c(v').
\end{equation}

\noindent\textbf{Hard match for real tuples.} We proceed in an analogous fashion as
for the {\em hard match for discrete tuples}, that is, we replace the
label $\ell(v)$ in Equation~\ref{eq:ksub} with the labeling procedure
$\Label^v$. In this case, however, we combine the real valued tuple of
corresponding vertices with the standard product as in Equation
\ref{eq:soft_real}:
\begin{equation}
  \kappa_{tuple}(v,v')= \sum_c prop_c(v) \cdot prop_c(v').
\end{equation}

\noindent\textbf{Soft match for mixed discrete and real tuples.} When dealing with
tuples of mixed discrete and real values, the contribution of the
kernels on the separate collections of discrete and real attributes
are combined via summation:
\begin{equation}
  \kappa_{tuple}(v,v')= \sum_d \mathbf{1}_{prop_d(v)=prop_d(v')} + \sum_c prop_c(v) \cdot prop_c(v')
\end{equation}
where indices $d$ and $c$ run exclusively over the discrete and
continous properties respectively.

\noindent\textbf{Hard match for mixed discrete and real tuples.} In an analogous
fashion, provided that $\ell(v)$ in Equation \ref{eq:ksub} is replaced
with the labeling procedure $\Label^v$ (see
Section~\ref{sec:exact_match}) we have:
\begin{equation}
  \kappa_{tuple}(v,v')= \prod_d \mathbf{1}_{prop_d(v)=prop_d(v')} + \sum_c prop_c(v) \cdot prop_c(v').
\end{equation}
In this way, each vertex receives a canonical label that
uniquely identifies it in the neighborhood graph. The discrete
labels of corresponding vertices are concatenated and matched for identity,
while the real tuples of corresponding vertices are combined via the standard
dot product.

\subsubsection{Domain knowledge bias via kernel points}
At times it is convenient, for efficiency reasons or to inject domain
knowledge into the kernel, to be able to explicitly select the
neighborhood subgraphs. We provide a way to do so, declaratively, by
introducing the set of {\em kernel points}, a subset of $V(G)$ which
includes all vertices associated with ground atoms of some specially
marked signatures.  We then redefine the relation $R_{r,d}(A,B,G)$
used in Equation~\ref{eq:kappard} like in Section~\ref{sec:nspdk} but
with the additional constraints that the roots of $A$ and $B$ be
kernel points.

Kernel points are typically vertices 
that are believed to represent information of high importance for the
task at hand. Vertices that are not kernel points contribute to the
kernel computation only when they occur in the neighborhoods of kernel points. 
In kLog, kernel points are declared as a list of domain
relations: all vertices that correspond to ground atoms of these
relations become kernel points.

\subsection{Viewpoints}
\label{sec:viewpoints}
The above approach effectively defines a kernel over interpretations
$$
K(z,z') = K(G_z,G_{z'})
$$
where $G_z$ is the result of graphicalization applied to
interpretation $z$.  For learning jobs such as classification or
regression on interpretations (see Table~\ref{tab:jobs}), this kernel
is directly usable in conjunction with plain kernel machines like
SVM. When moving to more complex jobs involving, e.g., classification
of entities or tuples of entities, the kernel induces a feature vector
$\phi(x,y)$ suitable for the application of a structured output
technique where $f(x)=\mathrm{arg}\max_y w\transpose\phi(x,y)$. Alternatively,
we may convert the structured output problem into a set of independent
subproblems as follows. For simplicity, assume the learning job
consists of a single relation $r$ of relational arity $n$. We call
each ground atom of $r$ a \textit{case}. Intuitively, cases correspond
to training targets or prediction-time queries in supervised learning.
Usually an interpretation contains several cases corresponding to
specific entities such as individual Web pages (as in
Section~\ref{ex:entity-classification}) or movies (as in
Section~\ref{ex:one-interpretation}), or tuples of entities for link
prediction problems (as in Section~\ref{sec:running-example}).  Given a
case $c\in y$, the \textit{viewpoint} of $c$, $W_c$, is the set of
vertices that are adjacent to $c$ in the graph.
\begin{figure}
  \centering
  \ifarxiv
  \includegraphics[width=0.5\textwidth]{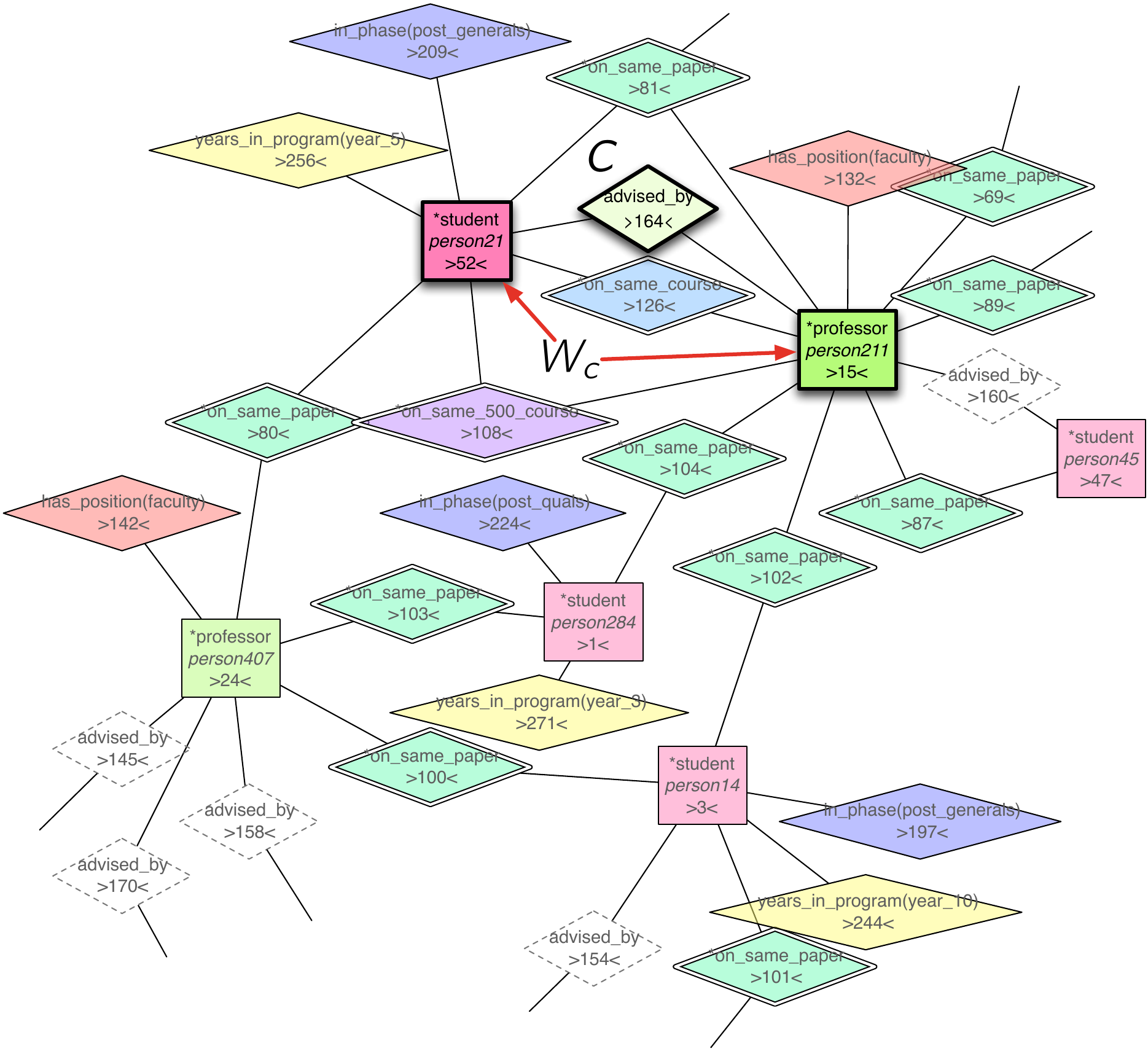}
  \else
  \includegraphics[width=0.5\textwidth]{Figures/viewpoints}
  \fi
  \caption{\label{fig:uwcse_fragment}
     A sample viewpoint ($W_c$) in the UW-CSE domain, and the corresponding mutilated graph where removed vertices are grayed out and the target case $c$ highlighted.}
\end{figure}
In order to define the kernel, we first consider the mutilated graph
$G_c$ where all vertices in $y$, except $c$, are removed (see
Figure~\ref{fig:uwcse_fragment} for an illustration).  We then define
a kernel $\hat{\kappa}$ on mutilated graphs, following the same
approach of the NSPDK, but with the {\em additional constraint} that the
first endpoint must be in $W_c$. The decomposition is thus defined as
$$
\hat{R}_{r,d} = \{(A,B,G_c) : N_r^v,N_r^u,v\in W_c,d^\star(u,v)=d\}.
$$
We obtain in this way a kernel ``centered'' around case $c$:
$$
\hat{K}(G_c,G'_{c'})=\sum_{r,d} \!\! \sum_{
  \begin{scriptsize} 
    \begin{array}{c} 
      A,B \in \hat{R}_{r,d}^{-1}(G_c)\\
      A',B' \in \hat{R}_{r,d}^{-1}(G'_{c'})
    \end{array} 
  \end{scriptsize} }\!\!\!\!\!\!\!\!\!\!\!\!
\delta(A,A')\delta(B,B')
$$
and finally we let
$$ K(G,G') = \sum_{c\in y,c'\in y'} \hat{K}(G_c,G'_{c'}).$$
This kernel corresponds to the potential 
$$
F(x,y) = w\transpose\sum_{c} \hat{\phi}(x,c)
$$
which is clearly maximized by maximizing, independently, all
sub-potentials $w\transpose\hat{\phi}(x,c)$ with respect to $c$.

By following this approach, we do not obtain a collective prediction
(individual ground atoms are predicted independently). Still, even in
this reduced setting, the kLog framework can be exploited in
conjunction with meta-learning approaches that surrogate collective
prediction. For example, Prolog predicates in intensional signatures
can effectively be used as expressive relational templates for stacked
graphical models~\cite{kou2007stacked} where input features for one
instance are computed from predictions on other related
instances. Results in Section~\ref{sec:uwcse} for the ``partial
information'' setting are obtained using a special form of stacking.

\subsection{Parameterization in kLog}
The kernels presented in this section, together with the
graphicalization procedure, yield a statistical model working in a
potentially very high-dimensional feature space. Although large-margin
learners offer some robustness against high-dimensional
representations, it is still the user's responsibility to choose appropriate
kernel parameters (\maxradius\ and \maxdistance) to avoid overfitting. It should
be noted that subgraphs in the NSPDK effectively act like templates
which are matched against graphicalized interpretations for new
data. Since identifiers themselves do not appear in the result of the
graphicalization procedure, the same template can match multiple times
realizing a form of parameter tying as in other statistical relational
learning methods.

\section{kLog in practice}
\label{sec:klog-in-practice}
In this section, we illustrate the use of kLog in a number of
application domains. All experimental results reported in
this section were obtained using LibSVM in binary classification,
multiclass, or regression mode, as appropriate.

\subsection{Predicting a single property of one interpretation}
\label{sec:molecules}
We now expand the ideas outlined in Example~\ref{ex:iid}.  Predicting
the biological activity of small molecules is a major task in
chemoinformatics and can help drug development \citep{Waterbeemd2003}
and toxicology \citep{Helma2003,Helma2005:book}. Most existing graph
kernels have been tested on data sets of small molecules (see,
e.g.,~\cite{horvath04,ralaivola2005graph,mahe2005gkm,Vishwanathan2010:jrnl,ceroni2007classification}). From
the kLog perspective the data consists of several interpretations, one
for each molecule. In the case of binary classification (e.g., active
vs.\ nonactive), there is a single target predicate whose truth state
corresponds to the class of the molecule. To evaluate kLog we used two
data sets.  The Bursi data
set~\cite{Kazius:2005:Derivation-and-validation-of-toxicophores}
consists of 4,337 molecular structures with associated mutagenicity
labels (2,401 mutagens and 1,936 nonmutagens) obtained from a
short-term in vitro assay that detects genetic damage.  The
Biodegradability data set~\cite{BlockeelEtAl:04} contains 328
compounds and the regression task is to predict their half-life for
aerobic aqueous biodegradation starting from molecular structure
and global molecular measurements.

\renewcommand{\arraystretch}{0.93}
\begin{table}
  \caption{\label{tab:bursiauroc}
    Results (AUROC, 10-fold cross-validation) on the Bursi data set with and without
    functional groups. The following kernel and SVM hyperparameters are varied: maximum radius \maxradius,
    maximum distance \maxdistance, and regularization parameter $C$.}
\centering
\begin{tabular}{ccccc}
$C$ & \maxdistance & \maxradius & With functional groups & Atom bond\tabularnewline
\hline 
\multirow{9}{*}{0.5} & \multirow{3}{*}{8} & 4 & $0.90\pm0.01$ & $0.88\pm0.01$\tabularnewline
 &  & 6 & $0.90\pm0.02$ & $0.88\pm0.02$\tabularnewline
 &  & 8 & $0.89\pm0.01$ & $0.88\pm0.01$\tabularnewline
\cline{2-5} 
 & \multirow{3}{*}{10} & 4 & $0.90\pm0.02$ & $0.88\pm0.02$\tabularnewline
 &  & 6 & $0.90\pm0.01$ & $0.88\pm0.01$\tabularnewline
 &  & 8 & $0.89\pm0.02$ & $0.88\pm0.02$\tabularnewline
\cline{2-5} 
 & \multirow{3}{*}{12} & 4 & $0.90\pm0.01$ & $0.88\pm0.01$\tabularnewline
 &  & 6 & $0.90\pm0.01$ & $0.88\pm0.01$\tabularnewline
 &  & 8 & $0.89\pm0.02$ & $0.88\pm0.02$\tabularnewline
\hline 
\multirow{9}{*}{1.0} & \multirow{3}{*}{8} & 4 & $0.91\pm0.01$ & $0.89\pm0.01$\tabularnewline
 &  & 6 & $0.91\pm0.02$ & $0.89\pm0.02$\tabularnewline
 &  & 8 & $0.90\pm0.01$ & $0.89\pm0.01$\tabularnewline
\cline{2-5} 
 & \multirow{3}{*}{10} & 4 & $0.91\pm0.02$ & $0.89\pm0.02$\tabularnewline
 &  & 6 & $0.91\pm0.02$ & $0.89\pm0.02$\tabularnewline
 &  & 8 & $0.90\pm0.02$ & $0.89\pm0.02$\tabularnewline
\cline{2-5} 
 & \multirow{3}{*}{12} & 4 & $0.91\pm0.01$ & $0.89\pm0.01$\tabularnewline
 &  & 6 & $0.91\pm0.02$ & $0.89\pm0.02$\tabularnewline
 &  & 8 & $0.90\pm0.01$ & $0.89\pm0.01$\tabularnewline
\hline 
\multirow{9}{*}{2.0} & \multirow{3}{*}{8} & 4 & $0.91\pm0.01$ & $0.90\pm0.01$\tabularnewline
 &  & 6 & $0.91\pm0.02$ & $0.90\pm0.02$\tabularnewline
 &  & 8 & $0.90\pm0.01$ & $0.89\pm0.01$\tabularnewline
\cline{2-5} 
 & \multirow{3}{*}{10} & 4 & $0.92\pm0.01$ & $0.90\pm0.01$\tabularnewline
 &  & 6 & $0.91\pm0.01$ & $0.90\pm0.01$\tabularnewline
 &  & 8 & $0.90\pm0.01$ & $0.89\pm0.01$\tabularnewline
\cline{2-5} 
 & \multirow{3}{*}{12} & 4 & $0.91\pm0.01$ & $0.89\pm0.01$\tabularnewline
 &  & 6 & $0.91\pm0.01$ & $0.90\pm0.01$\tabularnewline
 &  & 8 & $0.90\pm0.01$ & $0.89\pm0.01$\tabularnewline
\hline 
\multirow{9}{*}{5.0} & \multirow{3}{*}{8} & 4 & $0.91\pm0.01$ & $0.90\pm0.01$\tabularnewline
 &  & 6 & $0.91\pm0.02$ & $0.90\pm0.02$\tabularnewline
 &  & 8 & $0.91\pm0.02$ & $0.89\pm0.02$\tabularnewline
\cline{2-5} 
 & \multirow{3}{*}{10} & 4 & $0.91\pm0.01$ & $0.90\pm0.01$\tabularnewline
 &  & 6 & $0.91\pm0.01$ & $0.90\pm0.01$\tabularnewline
 &  & 8 & $0.90\pm0.01$ & $0.89\pm0.01$\tabularnewline
\cline{2-5} 
 & \multirow{3}{*}{12} & 4 & $0.91\pm0.02$ & $0.90\pm0.02$\tabularnewline
 &  & 6 & $0.91\pm0.02$ & $0.90\pm0.02$\tabularnewline
 &  & 8 & $0.90\pm0.01$ & $0.89\pm0.01$\tabularnewline
\end{tabular}
\end{table}


\begin{listing}
\begin{Verbatim}[fontsize=\footnotesize,fontfamily=helvetica,frame=single,numbers=left,numbersep=3pt,numberblanklines=false,commandchars=\\\{\}]
\textit{begin_domain}.
\textit{signature} \textbf{atm}(atom_id::\textit{self}, element::\textit{property})::\textit{intensional}.
atm(Atom, Element) :- a(Atom,Element), \textbackslash+(Element=h).

\textit{signature} \textbf{bnd}(atom_1@b::atm, atom_2@b::atm, type::\textit{property})::\textit{intensional}.
bnd(Atom1,Atom2,Type) :-
	b(Atom1,Atom2,NType), describeBondType(NType,Type),
	atm(Atom1,_), atm(Atom2,_).

\textit{signature} \textbf{fgroup}(fgroup_id::\textit{self}, group_type::\textit{property})::\textit{intensional}.
fgroup(Fg,Type) :- sub(Fg,Type,_).

\textit{signature} \textbf{fgmember}(fg::fgroup, atom::atm)::\textit{intensional}.
fgmember(Fg,Atom):- subat(Fg,Atom,_), atm(Atom,_).

\textit{signature} \textbf{fg_fused}(fg1@nil::fgroup, fg2@nil::fgroup, nrAtoms::\textit{property})::\textit{intensional}.
fg_fused(Fg1,Fg2,NrAtoms):- fused(Fg1,Fg2,AtomList), length(AtomList,NrAtoms).

\textit{signature} \textbf{fg_connected}(fg1@nil::fgroup, fg2@nil::fgroup,
					bondType::\textit{property})::\textit{intensional}.
fg_connected(Fg1,Fg2,BondType):-
	connected(Fg1,Fg2,Type,_AtomList),describeBondType(Type,BondType).

\textit{signature} \textbf{fg_linked}(fg::fgroup, alichain::fgroup, saturation::\textit{property})::\textit{intensional}.
fg_linked(FG,AliChain,Sat) :-
	linked(AliChain,Links,_BranchesEnds,Saturation),
	( Saturation = saturated -> Sat = saturated ; Sat = unsaturated ),
	member(link(FG,_A1,_A2),Links).

\textit{signature} \textbf{mutagenic}::\textit{extensional}.
\textit{end_domain}.
\end{Verbatim}
\caption{\label{lis:bursi}
  Complete specification of the Bursi domain.}
\end{listing}

Listing~\ref{lis:bursi} shows a kLog script for the Bursi domain.
Relevant predicates in the
extensional database are \textsf{a/2}, \textsf{b/3} (atoms and bonds,
respectively, extracted from the chemical structure), \textsf{sub/3}
(functional groups, computed by DMax Chemistry Assistant
\cite{Ando06:jrnl,DeGrave2010:jrnl}), \textsf{fused/3}, \textsf{connected/4}
(direct connection between two functional groups), \textsf{linked/4}
(connection between functional groups via an aliphatic
chain). Aromaticity (used in the bond-type property of \textsf{b/3})
was also computed by DMax Chemistry Assistant.  The intensional
signatures essentially serve the purpose of simplifying the original
data representation. For example \textsf{atm/2} omits some entities
(hydrogen atoms), and \textsf{fg\_fused/3} replaces a list of atoms by
its length. 
In signature \textsf{bnd(atom\_1@b::atm,atom\_2@b::atm)} we use a \textit{role
field} \textsf{b} (introduced by the symbol \textsf{@}). Using the same role
twice declares that the two atoms play the same
role in the chemical bond, i.e. that the relation is symmetric. In this way,
each bond can be represented by one tuple only, while a more
traditional relational representation, which is directional, would
require two tuples.
While this may at first sight appear to be only syntactic sugar, it
does provide extra abilities for modeling which is important in some
domains.  For instance, when modeling ring-structures in molecules,
traditional logical and relational learning systems need to employ
either lists to capture all the elements in a ring structure, or else
need to include all permutations of the atoms participating in a ring
structure. 
For rings involving 6 atoms,
this requires 6!=720 different tuples, an unnecessary blow-up. Also,
working with lists typically leads to complications such as having to
deal with a potentially infinite number of terms.
The target relation \textsf{mutagenic} has relational
arity zero since it is a property of the whole interpretation. 

As shown in Table~\ref{tab:bursiauroc}, results are relatively stable
with respect to the choice of kernel hyperparameter (maximum radius
and distance) and SVM regularization and essentially match the best
results reported in~\cite{Costa::Fast-neighborhood-subgraph} (AUROC
$0.92\pm0.02$) even without composition with a polynomial
kernel. These results are not surprising since the graphs generated by
kLog are very similar in this case to the expanded molecular graphs
used in~\cite{Costa::Fast-neighborhood-subgraph}.

We compared kLog to Tilde on the same task.
A kLog domain specification can be trivially ported to background 
knowledge for Tilde.  Both systems can then access the same set of Prolog atoms.
Table~\ref{tab:tilde-bursi} summarizes the results and can be compared with Table~\ref{tab:bursiauroc}.
Augmenting the language with the functional groups from \cite{DeGrave2010:jrnl} unexpectedly gave worse results 
in Tilde compared to a plain atom-bond language.  The presence of some functional groups correlates well with 
the target and hence those tests are used near the root of the decision tree.
Unfortunately, the greedy learner is unable to refine its hypothesis down to the atom level 
and relies almost exclusively on the coarse-grained functional groups.  
Such local-optimum traps can sometimes be escaped from by looking ahead further, 
but this is expensive and Tilde ran out of memory for lookaheads higher than 1.
Tilde's built-in bagging can boost its results, especially when functional groups are used.
A comparison between Tables~\ref{tab:tilde-bursi} and~\ref{tab:bursiauroc} shows
that Tilde with our language bias definition is not well suited for this problem.
\begin{table}
  \centering
  \caption{Tilde results on Bursi.}
  \label{tab:tilde-bursi}
\begin{tabular}{llll}
Setting & Model & AUC \\
\hline
Functional groups & Tilde             & 0.63 $\pm$ 0.09 \\
Functional groups & Bagging 20x Tilde & 0.79 $\pm$ 0.06 \\
Atom bonds & Tilde             & 0.80 $\pm$ 0.02  \\
Atom bonds & Bagging 20x Tilde & 0.83 $\pm$ 0.02 \\
\end{tabular}
\end{table}


The kLog code for biodegradability is similar to Bursi but being a regression 
task we have a target relation declared as
\begin{center}
  \textsf{signature \textbf{biodegradation}(\textit{halflife::property})::extensional.}
\end{center}
We estimated prediction performance by repeating five times a ten-fold
cross validation procedure as described in~\cite{BlockeelEtAl:04}
(using exactly the same folds in each trial).  Results --- rooted mean
squared error (RMSE), squared correlation coefficient (SCC), and mean
absolute percentage error (MAPE) --- are reported in
Table~\ref{tab:biodeg}.  For comparison, the best RMSE obtained by
kFOIL on this data set (and same folds) is 1.14 $\pm$ 0.04 (kFOIL was
shown to outperform Tilde and S-CART
in~\cite{Landwehr:2010:Fast-learning-of-relational}).

\begin{table}
  \centering
  \caption{kLog results on biodegradability.}
  \label{tab:biodeg}
\begin{tabular}{llll}
Setting & RMSE & SCC & MAPE \\
\hline
Functional groups & 1.07 $\pm$ 0.01 & 0.54 $\pm$ 0.01 & 14.01 $\pm$ 0.08 \\
Atom bonds & 1.13 $\pm$ 0.01 & 0.48 $\pm$ 0.01 & 14.55 $\pm$ 0.12 \\
\end{tabular}
\end{table}

\subsection{Link prediction}
\label{sec:uwcse}

\begin{figure}[p]
  \centering
  \ifarxiv
  \includegraphics[width=0.95\textwidth]{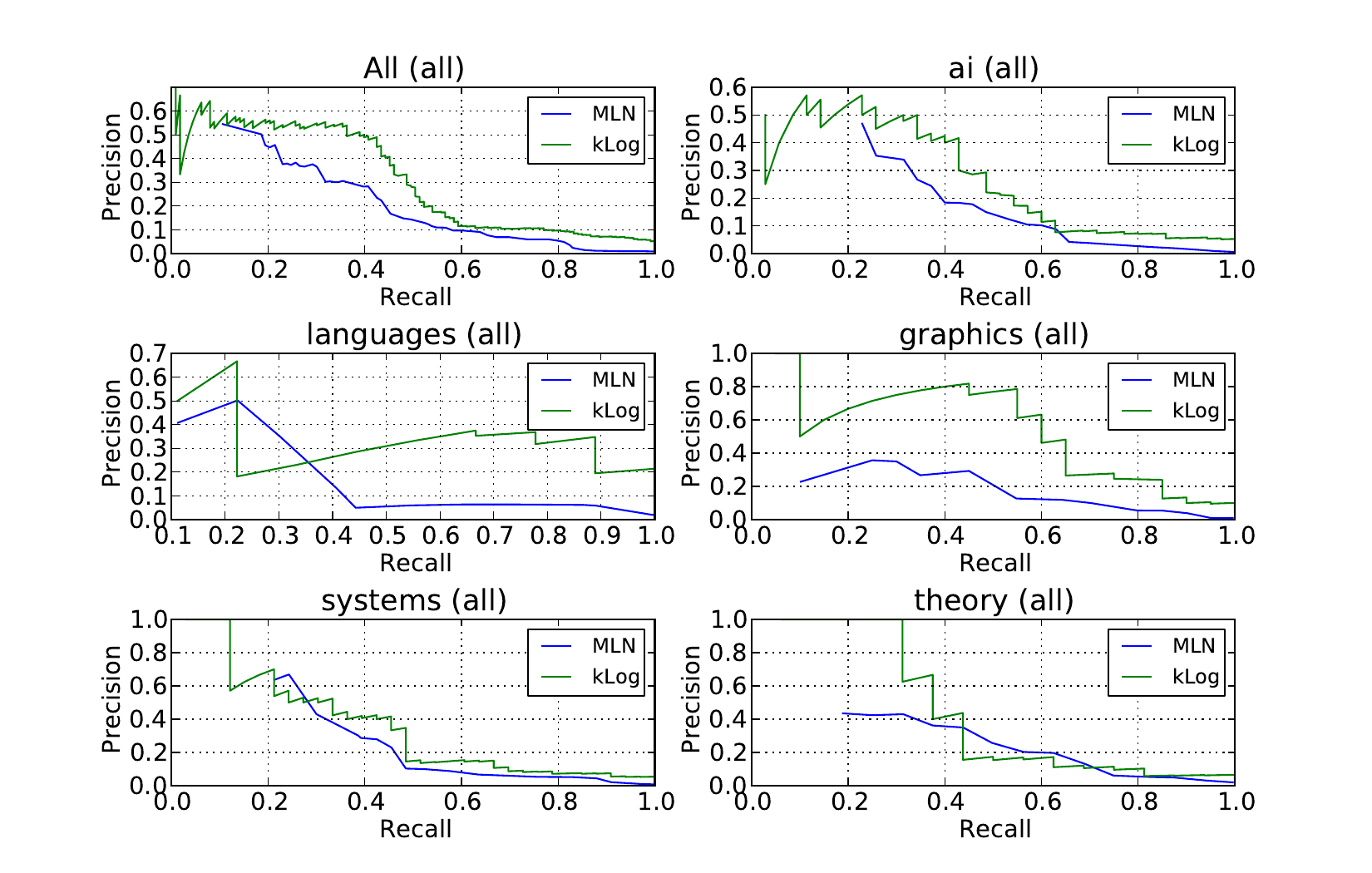}
  \else
  \includegraphics[width=0.95\textwidth]{Figures/rp-figure-all}
  \fi
  \caption{\label{fig:uwcse}
    Comparison between kLog and MLN on the UW-CSE domain (all information).}
\end{figure}

We report here experimental results on the UW-CSE domain that we have
already extensively described in Section~\ref{sec:running-example}.
To assess kLog behavior we evaluated prediction accuracy according to
the leave-one-research-group-out setup
of~\cite{Richardson:2006:Markov-logic-networks}, using the domain description of Listings~\ref{lis:signatures} and \ref{lis:intensional}, 
 together with a NSPDK kernel with distance 2,
radius 2, and soft match. Comparative results with respect to Markov
logic are reported in Figure~\ref{fig:uwcse} (MLN results published
in~\cite{Richardson:2006:Markov-logic-networks}). The whole 5-fold
procedure runs in about 20 seconds on a single core of a 2.5GHz Core i7 CPU. Compared
to MLNs, kLog in the current implementation has the disadvantage of
not performing collective assignment but the advantage of defining
more powerful features thanks to the graph kernel.  Additionally, MLN
results use a much larger knowledge base.  The advantage of kLog over
MLN in Figure~\ref{fig:uwcse} is due to the more powerful feature
space.  Indeed, when setting the graph kernel distance and radius to 0
and 1, respectively, the feature space has just one feature for each
ground signature, in close analogy to MLN. The empirical
performance (area under recall-precision curve, AURPC) of kLog using
the same set of signatures drops dramatically from 0.28 (distance 2, radius 2) to 0.09 (distance 1, radius 0). 

\begin{figure}[p]
  \centering
  \ifarxiv
  \includegraphics[width=0.95\textwidth]{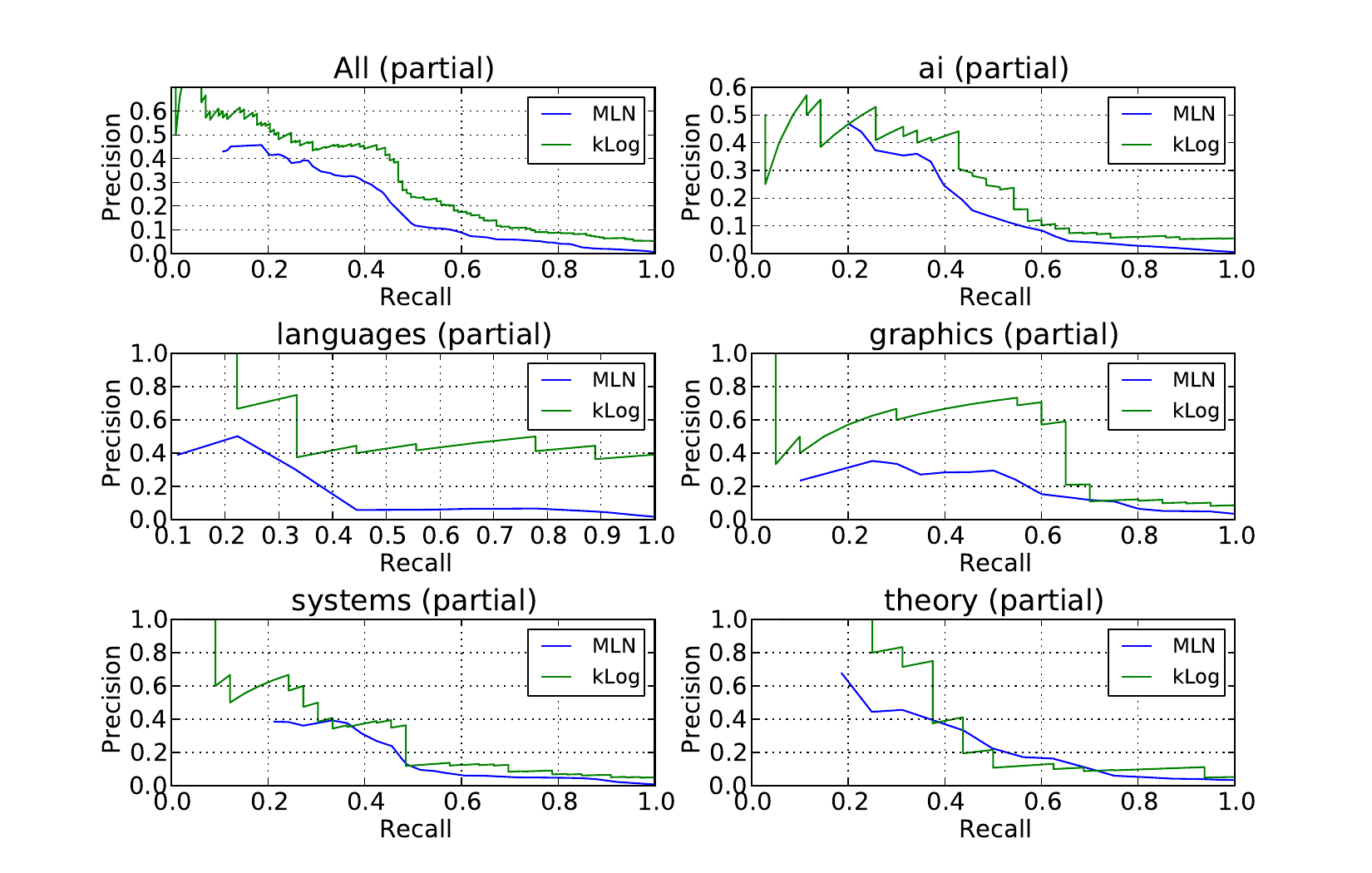}
  \else
  \includegraphics[width=0.95\textwidth]{Figures/rp-figure-partial}
  \fi
  \caption{\label{fig:uwcse-partial}
    Comparison between kLog and MLN on the UW-CSE domain (partial information).}
\end{figure}

In a second experiment, we predicted the relation \textsf{advised\_by}
starting from partial information (i.e., when relations
\textsf{Student} (and its complement \textsf{Professor}) are unknown,
as in~\cite{Richardson:2006:Markov-logic-networks}). In this case, we
created a pipeline of two predictors.  Our procedure is reminiscent of
stacked generalization~\cite{wolpert1992stacked}. In the first stage, a
leave-one-research-group-out cross-validation procedure was applied to the
training data to obtain predicted groundings for \textsf{Student} (a
binary classification task on entities). Predicted groundings were then
fed to the second stage which predicts the binary relation
\textsf{advised\_by}. The overall procedure was repeated using one
research group at the time for testing. Results are reported in
Figure~\ref{fig:uwcse-partial} (MLN results published
in~\cite{Richardson:2006:Markov-logic-networks}).

Since kLog is embedded in the programming language Prolog, it is easy
to use the output of one learning task as the input for the next one
as illustrated in the pipeline.  This is because both the inputs and
the outputs are relations. Relations are treated uniformly regardless
of whether they are defined intensionally, extensionally, or are the
result of a previous learning run.  Thus kLog satisfies what has been
called the {\em closure} principle in the context of inductive
databases
\cite{DBLP:journals/sigkdd/Raedt02,DBLP:reference/dmkdh/BoulicautJ10};
it is also this principle together with the embedding of kLog inside a
programming language (Prolog) that turns kLog into a true programming
language for machine learning
\cite{Mitchell06,DBLP:conf/ismis/RaedtN11,DBLP:conf/lrec/RizzoloR10}.
Such programming languages possess --- in addition to the usual
constructs --- also primitives for learning, that is, to specify the
inputs and the outputs of the learning problems.  In this way, they
support the development of software in which machine learning is
embedded without requiring the developer to be a machine learning
expert.  According to Mitchell \cite{Mitchell06}, the development 
of such languages is a long outstanding research question.


\subsection{Entity classification}
\begin{figure}
  \centering
  \ifarxiv
  \includegraphics[width=0.45\textwidth]{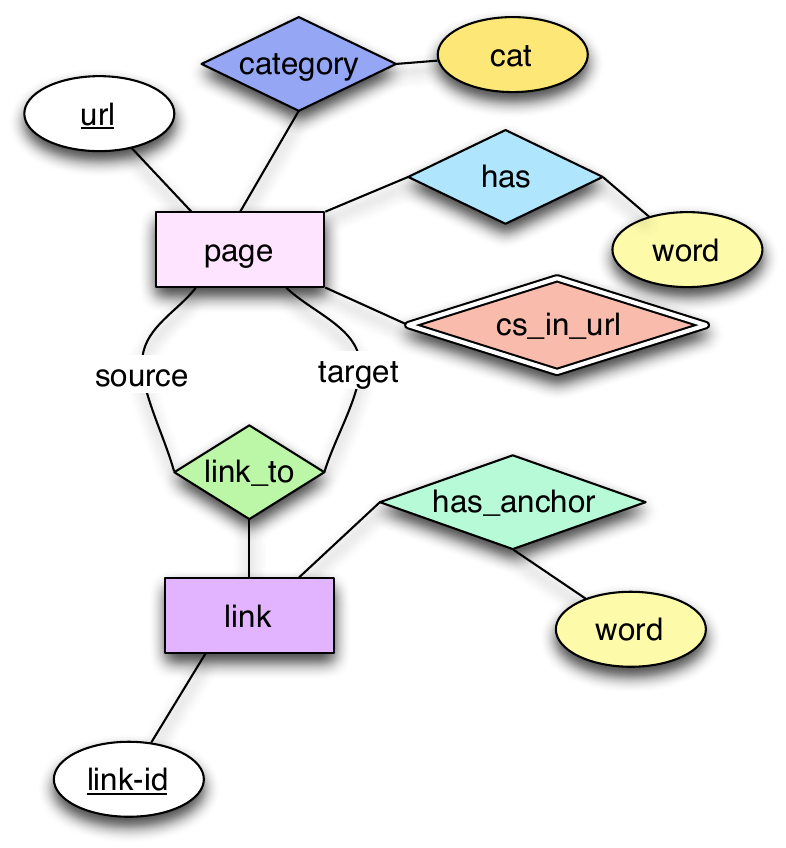}
  \else
  \includegraphics[width=0.45\textwidth]{Figures/WebKB-ER}
  \fi
  \caption{\label{fig:webkb}
     WebKB domain.}
\end{figure}
The WebKB data set~\cite{Craven:1998:Learning-to-Extract-Symbolic} has
been widely used to evaluate relational methods for text
categorization. It consists of academic Web pages from four computer
science departments and the task is to identify the category (such as
student page, course page, professor page, etc).
Figure~\ref{fig:webkb} shows the E/R diagram used in kLog.  One of the
most important relationships in this domain is \textnormal{\textsf{has}}, that
associates words to web pages. After graphicalization, vertices
representing webpages have large degree (at least the number of
words), making the standard NSPDK
of~\cite{Costa::Fast-neighborhood-subgraph} totally inadequate: even
by setting the maximum distance $\maxdistance=1$ and the maximum radius $\maxradius=2$,
the hard match would essentially create a distinct feature for every
page.  In this domain we can therefore appreciate the flexibility of
the kernel defined in
Section~\ref{sec:kernel}. In particular, the soft match
kernel creates histograms of word occurrences in the page, which is
very similar to the bag-of-words (with counts) representation that is
commonly used in text categorization problems. The additional
signature \textnormal{\textsf{cs\_in\_url}} embodies common sense background
knowledge that many course web pages contains the string ``cs''
followed by some digits and is intensionally defined using a Prolog predicate that 
holds true when the regular expression :cs(e*)[0-9]+: matches the page URL.

Empirical results using only four Universities (Cornell, Texas,
Washington, Wisconsin) in the leave-one-university-out setup are
reported in Table~\ref{tab:webkb}.
\begin{table}[p]
  \centering
  \caption{Results on WebKB (kLog): contingency table, accuracy, precision, recall, and F$_1$ measure per class. Last row reports micro-averages.}
  \label{tab:webkb}
\begin{tabular}{lrrrr|rrrr}
 & research & faculty & course & student & A & P & R & F$_1$ \\
\hline
research & 59 & 11 & 4 & 15 & 0.94 & 0.66 & 0.70 & 0.68 \\
faculty & 9 & 125 & 2 & 50 & 0.91 & 0.67 & 0.82 & 0.74 \\
course & 0 & 0 & 233 & 0 & 0.99 & 1.00 & 0.95 & 0.98 \\
student & 16 & 17 & 5 & 493 & 0.90 & 0.93 & 0.88 & 0.91 \\
\hline
Average &  &  &  &  & 0.88 & 0.89 & 0.88 & 0.88 \\
\end{tabular}
\end{table}
\begin{table}[p]
  \centering
  \caption{Results on WebKB (Markov logic): contingency table, accuracy, precision, recall, and F$_1$ measure per class. Last row reports micro-averages.}
  \label{tab:webkbmln}
\begin{tabular}{lrrrrrrrr}
 & research & faculty & course & student & A & P & R & F$_1$ \\
\hline
research & 42 & 3 & 7 & 32 & 0.95 & 0.95 & 0.50 & 0.66 \\
faculty & 1 & 91 & 1 & 60 & 0.92 & 0.88 & 0.59 & 0.71 \\
course & 0 & 1 & 233 & 10 & 0.98 & 0.95 & 0.95 & 0.95 \\
student & 1 & 9 & 3 & 545 & 0.89 & 0.84 & 0.98 & 0.90 \\
\hline
Average &  &  &  &  & 0.88 & 0.91 & 0.76 & 0.81 \\
\end{tabular}
\end{table}
\begin{table}[p]
  \centering
  \caption{Results on WebKB (Tilde): contingency table, accuracy, precision, recall, and F$_1$ measure per class. Last row reports micro-averages.}
  \label{tab:webkbtilde}
\begin{tabular}{lrrrrrrrr}
 & research & faculty & course & student & A & P & R & F$\_{}1$ \\
\hline
research & 39 & 9 & 1 & 35 & 0.93 & 0.65 & 0.46 & 0.54 \\
faculty & 11 & 110 & 1 & 31 & 0.91 & 0.69 & 0.72 & 0.71 \\
course & 1 & 1 & 235 & 7 & 0.99 & 0.99 & 0.96 & 0.98 \\
student & 9 & 39 & 1 & 509 & 0.88 & 0.87 & 0.91 & 0.89 \\
\hline
Average &  &  &  &  & 0.86 & 0.80 & 0.76 & 0.78 \\
\end{tabular}
\end{table}
We compared kLog to MLN and to Tilde on the same
task. For MLN we used the Alchemy system and the following
set of formulae, which essentially encode the same domain knowledge
exploited in kLog:
\begin{Verbatim}[fontsize=\footnotesize,fontfamily=helvetica,frame=single,numbers=left,numbersep=3pt,numberblanklines=false,commandchars=\\\{\}]
Has(+w,p) => Topic(+t,p)
!Has(+w,p) => Topic(+t,p)
CsInUrl(p) => Topic(+t,p)
!CsInUrl(p) => Topic(+t,p)
Topic(+c1, p1) ^ (Exist a LinkTo(a, p1, p2)) => Topic(+c2, p2)
HasAnchor(+w,a) ^ LinkTo(a, p1, p2) => Topic(+c, p2)
\end{Verbatim}
Ground atoms for the predicate \textnormal{\textsf{LinkTo}} were
actually precalculated externally in Prolog (same code as for the
kLog's intensional signature) since regular expressions are not
available in Alchemy. We did not enforce mutually exclusive
categories since results tended to be worse.  For learning we used the
preconditioned scaled conjugate gradient approach described
in~\cite{lowd2007efficient} and we tried a wide range of values for
the learning rate and the number of iterations. The best results,
reported in Table~\ref{tab:webkb}, used the trick of averaging
MLN weights across all iterations as in~\cite{lowd2007efficient}.  MC-SAT inference was used during
prediction. In spite of the advantage of MLN for using a collective
inference approach, results are comparable to those obtained with kLog
(MLN tends to overpredict the class ``student'', resulting in a
slightly lower average $F_1$ measure, but accuracies are
identical). Thus the feature extracted by kLog using the graph kernel
are capable of capturing enough contextual information from the input
portion of the data to obviate the lack of collective inference.

In the case of Tilde, we used the following language bias:
\begin{Verbatim}[fontsize=\footnotesize,fontfamily=helvetica,frame=single,numbers=left,numbersep=3pt,numberblanklines=false,commandchars=\\\{\}]
rmode((has(+U,Word,+Page), Word = #Word)).
rmode((cs_in_url(+U,+Page,V), V = #V)).
rmode((link_to(+U,-Link,-From,+Page), has_anchor(+U,Word,Link), Word = #Word)).
\end{Verbatim}
Results are reported in Table~\ref{tab:webkbtilde} and are slightly
worse than those attained by kLog and MLN.

Although the accuracies of the three methods are essentially
comparable, their requirements in terms of CPU time are dramatically
different: using a single core of a second generation Intel Core i7, kLog took
36s, Alchemy 27,041s (for 100 iterations, at which the best accuracy
is attained), and Tilde: 5,259s.

\subsection{Domains with a single interpretation}
\label{sec:imdb}
The Internet Movie Database (IMDb) collects information about movies
and their cast, people, and companies working in the motion picture
industry. We focus on predicting, for each movie, whether its first
weekend box-office receipts are over US\$2 million, a learning task
previously defined
in~\cite{Neville:2003:Collective-classification-with,Macskassy:2003:A-Simple-Relational-Classifier}.
The learning setting defined so far (learning from independent
interpretations) is not directly applicable since train and test data
must necessarily occur within the same interpretation.  The notion of
\textit{slicing} in kLog allows us to overcome this difficulty.  A
\textit{slice system} is a partition of the true ground atoms in a
given interpretation: $z=\{z(i_1),\dots,z(i_n)\}$ where the disjoint
sets $z(j)$ are called \textit{slices} and the index set
$I=\{i_1,\dots,i_n\}$ is endowed with a total order $\preceq$.  For
example, a natural choice for $I$ in the IMDb domain is the set of
movie production years (e.g., $\{1996,\dots,2005\}$), where the index
associated with a ground atom of an entity such as \textsf{actor} is
the debut year.

In this way, given two disjoint subsets of $I$, $T$ and $S$, such that
$\max_{\preceq}(T) \preceq \min_{\preceq}(S)$, it is reasonable during
training to use for some index $t\in T \setminus
\{\min_{\preceq}(T)\}$ the set of ground atoms $\{x(i): i\in T \wedge
i\preceq t\}_{t} \cup \{y(i): i\in T \wedge i\prec t\}_{t}$ (where
$i\prec t$ iff $i\preceq t$ and $i\neq t$) as the input portion of the
data, and $\{y(t)\}_{t}$ as the output portion (targets). Similarly,
during testing we can for each $s \in S$ use the set of ground atoms
$\{x(i): i\in S \wedge i\preceq s\}_s \cup \{y(i): i\in S \wedge
i\prec s\}_s$ for predicting $\{y(s)\}_s$.

The kLog data set was created after downloading the whole database
from \url{http://www.imdb.com}. Adult movies, movies produced outside
the US, and movies with no opening weekend data were
discarded. Persons and companies with a single appearance in this
subset of movies were also discarded. The resulting data set is
summarized in Table~\ref{tab:imdb}. We modeled the domain in kLog
using extensional signatures for movies, persons (actors, producers,
directors), and companies (distributors, production companies, special
effects companies). We additionally included intensional signatures
counting, for each movie the number of companies involved also in
other blockbuster movies.  We sliced the data set according to
production year, and starting from year $y=1997$, we trained on the
frame $\{y-1,y-2\}$ and tested on the frame $\{y\}$. Results (area
under the ROC curve) are summarized in Table~\ref{tab:imdb}, together
with comparative results against MLN and Tilde. In all three cases we
used the same set of ground atoms for training and testing, but in the
case of MLN the Alchemy software does not differentiate
between evidence and query ground atoms of the same predicate. We
therefore introduced an extra predicate called
\textsf{PrevBlockbuster} to inject evidence for years $\{y-3,y-4,\dots\}$, together with the hard rule
\begin{Verbatim}[fontsize=\footnotesize,fontfamily=helvetica,frame=single,numbers=left,numbersep=3pt,numberblanklines=false,commandchars=\\\{\}]
PrevBlockbuster(m) <=> Blockbuster(m).
\end{Verbatim}
and used \textsf{Blockbuster} as the query predicate when training on
years $\{y-1,y-2\}$ and testing on year $y$. MLN rules were designed
to capture past performance of actors, directors, distribution
companies, etc. For example:
\begin{Verbatim}[fontsize=\footnotesize,fontfamily=helvetica,frame=single,numbers=left,numbersep=3pt,numberblanklines=false,commandchars=\\\{\}]
ActedIn(p,m1) ^ ActedIn(p,m2) ^ Blockbuster(m1) ^ m1 != m2 => Blockbuster(m2)
\end{Verbatim}
In the case of Tilde, we used as background knowledge exactly the same intensional signature definitions used in kLog.
The MLN was trained using the discriminative weight learning algorithm implemented in Alchemy.
MCMC (Gibbs sampling) was used during prediction to obtain probabilities for all query atoms.

\begin{table}
  \centering
  \caption{Results (AUC) on the IMDb data set. Years 1995 and 1996 were only used for training. Years before 1995 (not shown) are always included in the training data.}
  \label{tab:imdb}
\begin{tabular}{rrrrrr}
Year & \# Movies & \# Facts & kLog & MLN & Tilde \\
\hline
1995 & 74 & 2483 & --- & --- & --- \\
1996 & 223 & 6406 & --- & --- & --- \\
1997 & 311 & 8031 & 0.86 & 0.79 & 0.80 \\
1998 & 332 & 7822 & 0.93 & 0.85 & 0.88 \\
1999 & 348 & 7842 & 0.89 & 0.85 & 0.85 \\
2000 & 381 & 8531 & 0.96 & 0.86 & 0.93 \\
2001 & 363 & 8443 & 0.95 & 0.86 & 0.91 \\
2002 & 370 & 8691 & 0.93 & 0.87 & 0.89 \\
2003 & 343 & 7626 & 0.95 & 0.88 & 0.87 \\
2004 & 371 & 8850 & 0.95 & 0.87 & 0.87 \\
2005 & 388 & 9093 & 0.92 & 0.84 & 0.83 \\
\hline
All &  &  & 0.93 $\pm$ 0.03 & 0.85 $\pm$ 0.03 & 0.87 $\pm$ 0.04 \\
\end{tabular}
\end{table}
On this data set, Tilde was the fastest system, completing all training
and test phases in 220s, followed by kLog (1,394s) and Alchemy
(12,812s). However, the AUC obtained by kLog is consistently higher across all prediction years.

\section{Related work}
\label{sec:related}
As kLog is a language for logical and relational learning with kernels
it is related to work on inductive logic programming, to statistical
relational learning, to graph kernels, and to propositionalization.
We now discuss each of these lines of work and their relation to kLog.

First, the underlying representation of the data that kLog employs at
the first level
is very close to that of standard inductive logic programming systems
such as Progol \cite{Mug95:jrnl}, Aleph \cite{aleph}, and Tilde \cite{1998-blockeel-0} in the sense that the input is
essentially (a variation of) a Prolog program for specifying the data
and the background knowledge. Prolog 
allows us to encode essentially any program as background
knowledge.  The E/R model used in kLog
is related to the Probabilistic Entity Relationship models introduced
by Heckerman et al.\ in \cite{Heckerman:2007:SRL}. The signatures play a similar role as the notion of a declarative bias in 
inductive logic programming~\cite{De-Raedt:2008:Logical-and-relational-learning}.  The combined use of the E/R model and 
the graphicalization has provided us with a powerful tool for
visualizing both the structure of the data (the E/R diagram)
as well as specific cases (through their graphs). This has proven
to be very helpful when preparing datasets for kLog. 
On the other hand, due to the adoption of a database framework, kLog forbids functors in the signature relations (though functors can be used inside predicates needed to
compute these relations inside  the background knowledge).    This
contrasts with some  inductive logic programming systems such as
Progol \cite{Mug95:jrnl} and Aleph \cite{aleph}.

Second, kLog is related to many existing statistical relational
learning  systems such as Markov logic
\cite{Richardson:2006:Markov-logic-networks}, probabilistic similarity
logic \cite{Brocheler:2010:Probabilistic-similarity-logic},
probabilistic relational models \cite{friedman99:learn}, Bayesian logic
programs  \cite{Kersting06},
and ProbLog \cite{De-Raedt:2007:ProbLog:-A-probabilistic-Prolog} in
that the representations of the inputs and outputs are essentially the
same, that is, both in kLog and in statistical relational learning systems
inputs are partial interpretations which are completed by predictions.
What kLog and statistical relational learning techniques 
have in common is that they both construct (implicitly or explicitly) graphs 
representing the instances. For statistical relational learning methods 
such as  Markov logic  \cite{Richardson:2006:Markov-logic-networks},
probabilistic relational models  \cite{friedman99:learn}, and Bayesian
logic programs  \cite{Kersting06} the knowledge-based model construction process will
result in a graphical model (Bayesian or Markov network) for each instance
representing a class of probability distributions, while in
kLog the process of graphicalization results in a graph 
representing an instance by unrolling the  E/R-diagram.  
Statistical relational learning systems then learn a probability
distribution using the features and parameters in the graphical model, while kLog
learns a function using the features derived by the kernel from the
graphs. Notice that in both cases, the resulting features are tied together. 
Indeed, in statistical relational learning each ground instance 
of a particular template or expression that occurs in the graph has the same parameters.
kLog features correspond to subgraphs that represent relational templates and 
that may match (and hence be grounded) multiple times in the graphicalization.
As each such feature has a single weight, kLog also realizes parameter tying
in a similar way as statistical relational learning methods. 
One difference between these statistical relational learning models and
kLog is that the former do not really have a second level as does
kLog. Indeed, the knowledge base model construction process directly
generates the graphical model that includes all the features used for
learning, while in kLog these features are derived from the graph kernel.
While statistical relational learning systems have been commonly used for collective
learning, this is still a question for further research within kLog.
A combination of structured-output learning \cite{Tsochantaridis:2006:Large-margin-methods} and iterative approaches (as
incorporated in the EM algorithm) can form the basis for further work
in this direction.  
Another interesting but more speculative direction for future work is concerned with lifted inference. Lifted inference has 
been the focus of a lot of attention in statistical relational learning; see \cite{DBLP:conf/ecai/Kersting12} for an overview.   One view on lifted inference is that it is trying to exploit symmetries in 
the graphical models that would be the result of the knowledge based model construction step, e.g., \cite{Kersting:2009:CBP:1795114.1795147}. From this perspective, it might be interesting to explore the use of symmetries in graphs and features 
constructed by kLog.

kLog builds also upon the many results on learning with graph kernels, see
\cite{Gartner03} for an overview.  A distinguishing feature of kLog is,
however, that the graphs obtained by graphicalizing a relational representation contain very rich labels, which can be both symbolic
and numeric. This contrasts with the kind of graphs needed to
represent for instance small molecules. In this regard, kLog is close
in spirit to the work of
\cite{Wachman:2007:Learning-from-interpretations:},
who define a kernel on hypergraphs, where hypergraphs are used to
represent relational interpretations. 
A distinctive feature of kLog is automatic graphicalization of
relational representations, which also allows users to naturally specify
multitask and collective learning tasks.

The graphicalization approach introduced in kLog is closely related to
the notion of propositionalization, a commonly applied technique in
logical and relational learning
\cite{2001-kramer-0,De-Raedt:2008:Logical-and-relational-learning} to
generate features from a relational representation. The advantage of
graphicalization is that the obtained graphs are essentially
equivalent to the relational representation and that --- in contrast to
the existing propositionalization approaches in logical and relational
learning --- this does not result in a loss of information. 
After graphicalization, any graph kernel can in principle be applied to the resulting graphs. 
Even though many of these kernels (such as the one
used in kLog) compute --- implicitly or explicitly --- a feature
vector, the dimensionality of the obtained vector is far beyond that
employed by traditional propositionalization approaches.  kFOIL
\cite{Landwehr:2010:Fast-learning-of-relational} is one such
propositionalization technique that has been tightly integrated with a
kernel-based method. It greedily derives a (small) set of features in
a way that resembles the rule-learning algorithm of FOIL
\cite{Qui90-ML:jrnl}.  

Several other approaches to relational learning and mining have employed graph-based encodings of the relational data, e.g., \cite{Rossi:2012:TGD:2444851.2444861,lao2010relational,cook2006mining,sun2012mining}. 
kLog encodes a set of ground atoms into a bipartite undirected graph whose nodes are true ground atoms and whose edges connect an entity atom to a relationship atom if the identifier of the former appears as an argument in the latter.
This differs from the usual encoding employed in graph-based approaches to relational learning
and mining, which typically use labeled edges to directly represent the relationships between
the nodes corresponding to the entities.  Furthermore, these approaches typically use
the graph-based representation as the single representation, and unlike 
kLog do neither consider the graph-based representation as an intermediate representation
nor work at three levels of representation (logical, graph-based and feature-based).

Other domain specific languages for machine learning have been
developed whose goals are closely related to those of kLog. Learning
based Java~\cite{DBLP:conf/lrec/RizzoloR10} was designed to
specifically address applications in natural language processing. It
builds on the concept of \textit{data-driven compilation} to perform
feature extraction and nicely exploits the \textit{constrained
  conditional model} framework~\cite{Chang08} for structured output
learning. FACTORIE~\cite{McCallum09} allows users to concisely define
features used in a factor graph and, consequently, arbitrarily
connected conditional random fields. Like with MLN, there is an
immediate dependency of the feature space on the sentences of the
language, whereas in kLog this dependency is indirect since the exact
feature space is eventually defined by the graph kernel.

\section{Conclusions}
\label{sec:conclusions}

We have introduced a novel language for logical and relational
learning called kLog. It tightly integrates logical and relational
learning with kernel methods and constitutes a principled framework
for statistical relational learning based on kernel methods rather
than on graphical models.  kLog uses a representation that is based on
E/R modeling, which is close to representations being
used by contemporary statistical relational learners. kLog first
performs graphicalization, that is, it computes a set of labeled
graphs that are equivalent to the original representation, and then
employs a graph kernel to realize statistical learning.  We have shown
that the kLog framework can be used to formulate and address a wide
range of learning tasks, that it performs at least comparably to
state-of-the-art statistical relational learning techniques,
and also that it can be used as a programming language for machine
learning.

The system presented in this paper is a first step towards a kernel-based language for relational learning but there are unanswered questions and interesting open directions for further research. One important aspect is the possibility of performing collective classification (or structured output prediction). 
Structured output learning problems can be naturally defined within kLog's semantics: graphicalization followed by a graph kernel yields a \textit{joint} feature vector $\phi(x,y)$ where $y$ are the groundings of the output predicates. Collective prediction amounts to maximizing $w\transpose\phi(x,y)$ with respect to $y$. There are two important cases to consider. If groundings in $y$ do not affect the graph structure (because they never affect the intensional signatures) then the collective classification problem is not more complicated than in related SRL systems. For example, if dependencies have a regular sequential structure, dynamic programming can be used for this step, exactly as in conditional random fields (indeed, collective classification has been succesfully exploited within kLog in an application to natural language test segmentation~\cite{Verbeke2012b}). 
However, in general, changing $y$ during search will also change the graph structure. 
In principle it is feasible to redo graphicalization from scratch during search, apply the graph kernel again and eventually evaluate $w\transpose\phi(x,y)$, but such a naive approach would of course be very inefficient. 
Developing clever and faster algorithms for this purpose is an interesting open issue. It should be remarked, however, that even without collective classification, kLog achieves good empirical results thanks to the fact that features produced by the graph kernel provide a wide relational context. Better understanding of generalization for structured prediction models has begun to emerge (see~\cite{icml2013_london13} and references therein) and a theoretical analysis of learning within the present kLog setting is another potential direction for future research.


The graph kernel that is currently employed in kLog makes use of the
notion of topological distances to define the concept of
neighborhoods. In this way, given a predicate of interest, properties
of ``nearby'' tuples are combined to generate features relevant for
that predicate. As a consequence, when topological distances are not
informative (e.g., in the case of dense graphs with small diameter)
then large fractions of the graph become accessible to any
neighborhood and the features induced for a specific predicate cease
to be discriminative. In these cases (typical when dealing with
small-world networks), kernels with a different type of bias (e.g., 
flow-based kernels) are more appropriate. The implementation of a
library of kernels suitable for different types of graphs, as well as the integration of other existing graph kernels (such as~\cite{Gartner:2008:Kernels-for-str,Vishwanathan2010:jrnl,shervashidze2011weisfeiler}) in the kLog framework, is therefore
an important direction for future development.

Furthermore, even though kLog's current implementation is quite
performant, there are interesting implementation issues to be studied.
Many of these are similar to those employed in statistical relational
learning systems such as Alchemy
\cite{Richardson:2006:Markov-logic-networks} and ProbLog
\cite{DBLP:journals/tplp/KimmigDRCR11}.  

kLog is being actively used for developing applications. We are
currently exploring applications of kLog in natural language
processing~\cite{Verbeke2012,Verbeke2012b,Kordjamshidi2012} and in
computer vision~\cite{Antanas2012,AntanasHFTR13,Antanas2011}.

\subsection*{Acknowledgments}
We thank Tom Schrijvers for his dependency analysis code.  PF was
supported partially by KU Leuven SF/09/014 and partially by Italian
Ministry of University and Research PRIN 2009LNP494. KDG was supported
partially by KU Leuven GOA/08/008 and partially by ERC Starting Grant
240186 ``MiGraNT''. Finally, we would like to thank the anonymous
reviewers and the associate editor for their very useful and
constructive comments.

\input{appendix5.tex}
\bibliographystyle{elsarticle-num} 
\bibliography{k}
\end{document}

%% file: appendix5.tex
\appendix
\newpage
\section{Syntax of the kLog domain declaration section}
\label{sec:syntax_appendix}
A kLog program consists of Prolog code augmented by a domain
declaration section delimited by the pair of keywords
\textsf{begin\_domain} and \textsf{end\_domain} and one or more 
 \textit{signature declarations}.  A signature declaration
consists of a signature header followed by one or more Prolog
clauses. 
Clauses in a signature declaration
form the declaration of \textit{signature predicates} and are
automatically connected to the current signature header. There are a
few signature predicates with a special meaning for kLog, as discussed
in this section. 
A brief BNF description of the grammar of kLog
domains is given in Figure~\ref{fig:syntax}.

Additionally, kLog provides a library of Prolog predicates for
handling data, learning, and performance measurement.

\begin{figure}[th]
  \begin{grammar}
    [(colon){$\rightarrow$}]
    [(semicolon)$|$]
    [(comma){}]
    [(period){\\}]
    [(quote){\begin{bfseries}\begin{sffamily}}{\end{sffamily}\end{bfseries}}]
    [(nonterminal){\begin{itshape}}{\end{itshape}}]
    <domain> : "begin\_domain." <signatures> "end\_domain.".
    <signatures> : <signature> ; [<signature> <signatures>].
    <signature> : <header> [<sig\_clauses>].
    <sig\_clauses> : <sig\_clause> ; [<sig\_clause> <sig\_clauses>].
    <sig\_clause> : <Prolog\_clause>.
    <header> : "signature" <sig\_name> "(" <args> ")" "::" <level> ".".
    <sig\_name> : <Prolog\_atom>.
    <args> : <arg> ; [<arg> <args>].
    <arg> : <column\_name> [<role\_overrider>] "::" <type>.
    <column\_name> : <Prolog\_atom>.
    <role\_overrider> : "@" <role>.
    <role> : <Prolog\_atom>.
    <type> : "self" | <sig\_name> | "property".
    <level> : "intensional" ; "extensional".
  \end{grammar}
  \caption{\label{fig:syntax}kLog syntax}
\end{figure}

\section{Definitions}
\label{sec:definitions}
For the sake of completeness we report here a number of graph
theoretical definitions used in the paper.  We closely follow the
notation in \cite{GrossYellen2005}.  A graph $G=(V,E)$ consists of two
sets $V$ and $E$. The notation $V(G)$ and $E(G)$ is used when $G$ is
not the only graph considered. The elements of $V$ are called {\em
  vertices} and the elements of $E$ are called {\em edges}.  Each edge
has a set of two elements in $V$ associated with it, which are called
its {\em endpoints}, which we denote by concatenating the vertices
variables, e.g. we represent the edge between the vertices $u$ and $v$
with $uv$. In this case we say that $u$ is \textit{adjacent} to $v$ and that $uv$ is \textit{incident} on $u$ and $v$.
The {\em  degree} of a vertex is number of edges incident on it.
A graph is {\em bipartite} if its vertex set can be partitioned into
two subsets $X$ and $Y$ so that every edge has one end in $X$ and the
other in $Y$. 
A graph is {\em rooted} when we distinguish one of its
vertices, called {\em root}.
The {\em neighborhood} of a vertex $v$ is the set of vertices that are
adjacent to $v$ and is indicated with $N(v)$.  The {\em neighborhood}
of radius $r$ of a vertex $v$ is the set of vertices at a distance
less than or equal to $r$ from $v$ and is denoted by $N_r(v)$.  In a
graph $G$, the {\em induced subgraph} on a set of vertices
$W=\{w_1,\ldots,w_k\}$ is a graph that has $W$ as its vertex set and
it contains every edge of $G$ whose endpoints are in $W$.  
The {\em neighborhood subgraph} of radius $r$ of vertex $v$ is the
subgraph induced by the neighborhood of radius $r$ of $v$ and is
denoted by $\CalN^v_r$. A {\em labeled graph} is a graph whose
vertices and/or edges are labeled, possibly with repetitions, using
symbols from a finite alphabet. We denote the function that maps the
vertex/edge to the label symbol as $\ell$.  Two simple graphs
$G_1=(V_1,E_1)$ and $G_2=(V_2,E_2)$ are {\em isomorphic}, which we
denote by $G_1 \simeq G_2$, if there is a bijection $\phi : V_1
\rightarrow V_2$, such that for any two vertices $u,v \in V_1$, there
is an edge $uv$ if and only if there is an edge $\phi(u)\phi(v)$ in
$G_2$. An isomorphism is a structure-preserving bijection. Two labeled
graphs are isomorphic if there is an isomorphism that preserves also
the label information, i.e. $\ell(\phi(v))=\ell(v)$.  An {\em
  isomorphism invariant} or {\em graph invariant} is a graph property
that is identical for two isomorphic graphs (e.g. the number of
vertices and/or edges). A {\em certificate for isomorphism} is an
isomorphism invariant that is identical for two graphs if and only if
they are isomorphic.


\section{Decomposition kernels}
\label{sec:decomposition_kernels}
We follow the notation in \cite{Haussler99:other}. Given a set $X$
and a function $K : X \times X \rightarrow \RSet$, we say that $K$ is a
{\em kernel on} $X \times X$ if $K$ is symmetric, i.e. if for any $x$
and $y \in X$ $K(x,y)=K(y,x)$, and if $K$ is {\em
  positive-semidefinite}, i.e. if for any $N \ge 1$ and any
$x_1,\ldots,x_N \in X$, the matrix $K$ defined by $K_{ij}=K(x_i,x_j)$
is positive-semidefinite, that is $\sum_{ij}c_ic_jK_{ij} \ge 0$ for
all $c_1,\ldots,c_N \in \RSet$ or equivalently if all its eigenvalues
are nonnegative. It is easy to see that if each $x \in X$ can be
represented as $\phi(x)=\{\phi_n(x)\}_{n \ge 1}$ such that $K$ is the
ordinary $l_2$ dot product $K(x,y)= \langle \phi(x),\phi(y) \rangle
=\sum_n \phi_n(x)\phi_n(y)$ then $K$ is a kernel. The converse is also
true under reasonable assumptions (which are almost always verified)
on $X$ and $K$, that is, a given kernel $K$ can be represented as
$K(x,y)= \langle \phi(x),\phi(y) \rangle$ for some choice of
$\phi$. In particular it holds for any kernel $K$ over $X \times X$
where $X$ is a countable set. The vector space induced by $\phi$ is
called the {\em feature space}. Note that it follows from the
definition of positive-semidefinite that the {\em zero-extension} of a
kernel is a valid kernel, that is, if $S \subseteq X $ and $K$ is a
kernel on $S \times S$ then $K$ may be extended to be a kernel on $X
\times X$ by defining $K(x,y)=0$ if $x$ or $y$ is not in $S$.  It is
easy to show that kernels are closed under summation, i.e. a sum of kernels
is a valid kernel.

Let now $x \in X$ be a {\em composite structure} such that we can
define $x_1,\ldots,x_D$ as its parts\footnote{Note that the set of
  parts needs not be a partition for the composite structure, i.e. the
  parts may ``overlap''.}. Each part is such that $x_d \in X_d$ for
$d=1,\ldots,D$ with $D \ge 1$ where each $X_d$ is a countable set. Let
$R$ be the relation defined on the set $X_1 \times \ldots \times X_D
\times X$, such that $R(x_1,\ldots,x_D,x)$ is true iff
$x_1,\ldots,x_D$ are the parts of $x$. We denote with $R^{-1}(x)$ the
inverse relation that yields the parts of $x$, that is
$R^{-1}(x)=\{x_1,\ldots,x_D : R(x_1,\ldots,x_D,x)\}$.  In
\cite{Haussler99:other} it is demonstrated that, if there exist a
kernel $K_d$ over $X_d \times X_d$ for each $d=1,\ldots,D$, and if two
instances $x,y \in X$ can be decomposed in $x_1,\ldots,x_d$ and
$y_1,\ldots,y_d$, then the following generalized convolution:
\begin{equation}
K(x,y)=\!\!\!\!\!\sum_{
 \begin{scriptsize} 
   \begin{array}{c} 
     x_1,\ldots,x_m \in R^{-1}(x)\\
     y_1,\ldots,y_m \in R^{-1}(y)
 \end{array} 
 \end{scriptsize} }
\prod_{m=1}^MK_m(x_m,y_m)
\end{equation}

is a valid kernel called a {\em convolution} or {\em decomposition}
kernel\footnote{To be precise, the valid kernel is the zero-extension
  of $K$ to $X \times X$ since $R^{-1}(x)$ is not guaranteed to yield
  a non-empty set for all $x \in X$. }. In words: a decomposition
kernel is a sum (over all possible ways to decompose a structured
instance) of the product of valid kernels over the parts of the
instance.

\section{Graph invariant pseudo-identifier computation}
\label{sec:graphinvariant}

\begin{figure}
  \centering
  \ifarxiv
  \includegraphics[width=1\textwidth]{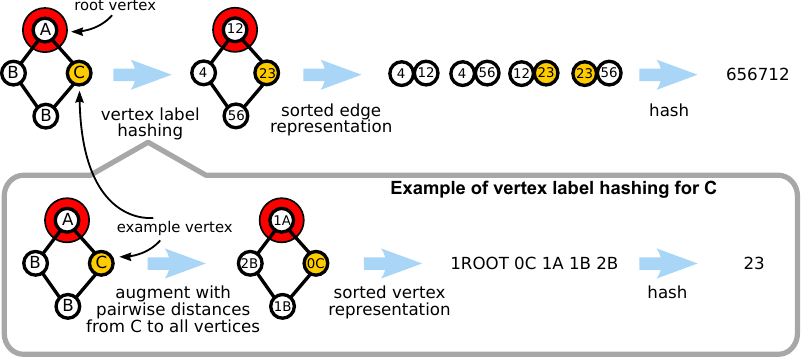}
  \else
  \includegraphics[width=1\textwidth]{Figures/graph_hashing.pdf}
  \fi
  \caption{\label{fig:nspdk_invar} Graph invariant computation: the graph
    hash is computed as the hash of the sorted list of edge hashes. An
    edge hash is computed as the hash of the sequence of the two
    endpoints hashes and the edge label. The endpoint hash is computed
    as the hash of the sorted list of distance-vertex label pairs.}
\end{figure}

We obtain the pseudo-identifier of a rooted neighborhood graph
$G_h$ by first constructing a graph invariant encoding $\Label^g(G_h)$. 
Then we apply a hash function $H(\Label^g(G_h)) \rightarrow \IN
$ to the encoding.  
The algorithm was first described in 
\cite{Costa::Fast-neighborhood-subgraph} but we present it again here for self-sufficiency. 
Figure \ref{fig:nspdk_invar} shows an overview.

Note that we cannot hope to exhibit an efficient
certificate for isomorphism in this way, and in general there can be
collisions between two non-isomorphic graphs, either because these are
assigned the same encoding or because the hashing procedure introduces
a collision even when the encodings are different.

Hashing is not a novel idea in machine learning; it is commonly used,
e.g., for creating compact representations of molecular
structures~\cite{ralaivola2005graph}, and has been advocated as a tool
for compressing very high-dimensional feature
spaces~\cite{shi2009hash}. In the present context, hashing is mainly
motivated by the computational efficiency gained by approximating the
isomorphism test.

The graph encoding $\Label^g(G_h)$ that we propose is best described by
introducing two new label functions: one for the vertices and one for
the edges, denoted $\Label^v$ and $\Label^e$ respectively. $\Label^v(v)$
assigns to vertex $v$ a lexicographically sorted sequence of pairs
composed by a topological distance and a vertex label, that is,
$\Label^v(v)$ returns a sorted list of pairs $\langle \CalD(v,u),
\ell(u) \rangle$ for all $u \in G_h$. Moreover, since $G_h$ is a
rooted graph, we can use the knowledge about the identity of the root
vertex $h$ and prepend to the returned list the additional information
of the distance from the root node $\CalD(v,h)$.
The new edge label is produced by composing the new vertex labels with
the original edge label, that is $\Label^e(uv)$ assigns to edge $uv$
the triplet $\langle \Label^v(u) ,\Label^v(v) , \ell(uv)
\rangle$. Finally $\Label^g(G_h)$ assigns to the rooted graph $G_h$
the lexicographically sorted list of $\Label^e(uv)$ for all $uv \in
E(G_h)$. In words: we relabel each vertex with a sequence that encodes
the vertex distance from all other (labeled) vertices (plus the
distance from the root vertex); the graph encoding is obtained as the
sorted edge list, where each edge is annotated with the endpoints' new
labels. For a proof that $\Label^g(G_h)$ is graph invariant, see 
\cite[p.\ 53]{degrave2011:phd}.

We finally resort to a Merkle-Damg{\aa}rd construction-based hashing
function for variable-length data to map the various lists to
integers, that is, we map the distance-label pairs, the new vertex
labels, the new edge labels and the new edge sequences to integers (in
this order). Note that it is trivial to control the size of the
feature space by choosing the hash codomain size (or alternatively the
bit size for the returned hashed values).  Naturally there is a
tradeoff between the size of the feature space and the number of hash
collisions.
